\documentclass{article}

\usepackage[T1]{fontenc}
\usepackage{newunicodechar}
\newunicodechar{’}{\textquoteright}
\usepackage{microtype}
\usepackage{graphicx}
\usepackage{subcaption}
\usepackage{caption}
\usepackage{booktabs}
\usepackage[dvipsnames]{xcolor}

\definecolor{cvprblue}{rgb}{0.21,0.49,0.74}
\usepackage[pagebackref,breaklinks,colorlinks,allcolors=cvprblue]{hyperref}

\PassOptionsToPackage{numbers,sort&compress}{natbib}
\usepackage[preprint]{icml2026}
\usepackage{amsmath}
\usepackage{amssymb}
\usepackage{mathtools}
\usepackage[most]{tcolorbox}
\tcbset{
  vhpromptbox/.style={
    colback=gray!4,
    colframe=black!65,
    boxrule=0.6pt,
    arc=2pt,
    left=6pt,
    right=6pt,
    top=5pt,
    bottom=5pt,
    fonttitle=\bfseries\small,
    fontupper=\footnotesize\ttfamily
  }
}
\usepackage[capitalize,noabbrev]{cleveref}
\usepackage{xurl}
\makeatletter
\newcommand{\layoutplaceholder}[2]{%
  \begingroup
  \setlength{\unitlength}{1pt}%
  \noindent
  \begin{picture}(\strip@pt\textwidth,#1)
    \put(0,0){\dashbox{3}(\strip@pt\textwidth,#1){%
      \parbox{0.9\textwidth}{\centering\normalfont\small #2}}}%
  \end{picture}%
  \endgroup
}
\makeatother


\newcommand{\vhfirstpageteaser}{%
  \par\vspace{2pt}%
  \noindent\makebox[\textwidth][c]{%
    \includegraphics[width=0.91\textwidth]{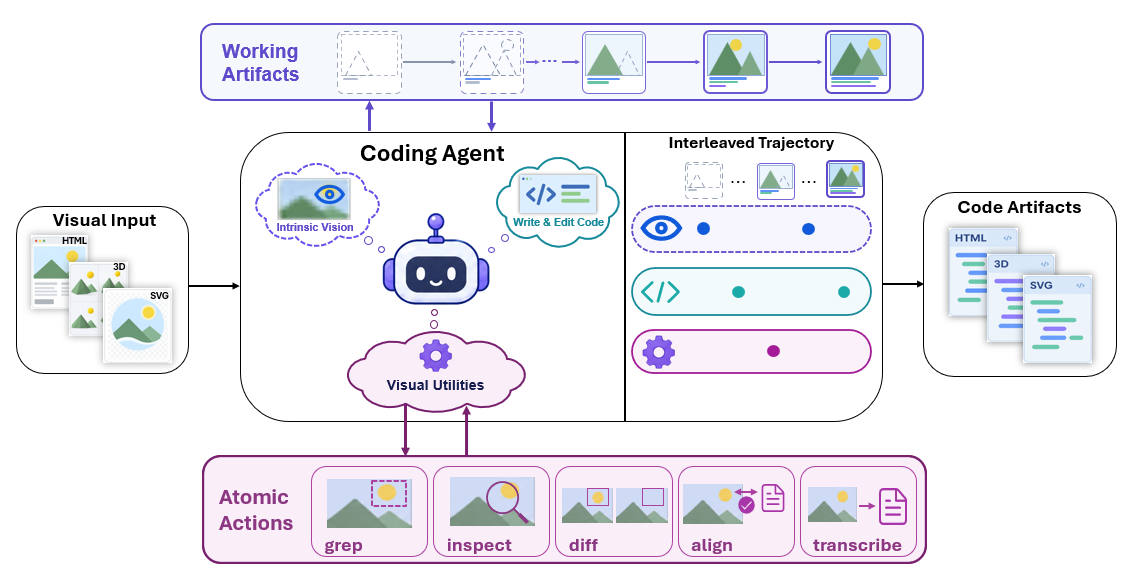}%
  }\par
  \captionof{figure}{\textbf{Overview of VHarness, an instantiation of the Vision-as-a-Tool (VaaT) paradigm.} Given a target visual input across Image-to-HTML, Image-to-SVG, and Image-to-3D tasks, a multimodal or text-only coding agent iteratively refines rendered \emph{Working Artifacts} toward the final \emph{Code Artifacts}. Rather than following a fixed workflow, the coding agent chooses among intrinsic vision when available (dashed blue), code writing and editing (green), and external visual utilities (purple) on demand, as shown in the \emph{Interleaved Trajectory}.}
  \label{fig:teaser}%
  \vspace{2pt}%
}

\newcommand{\vharchitecture}{%
  \begin{figure*}[!t]
    \centering
    \includegraphics[width=0.82\textwidth]{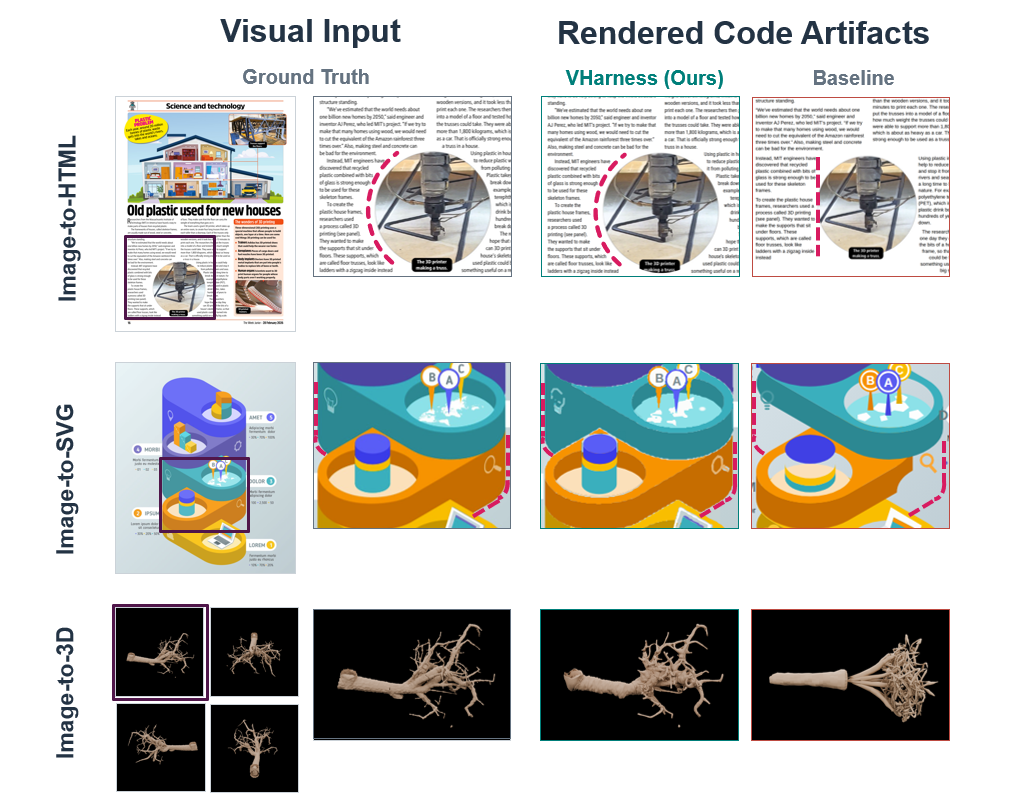}
    \caption{\textbf{Qualitative comparisons across Image-to-HTML, Image-to-SVG, and Image-to-3D tasks.} From left to right, we show the full reference input, the enlarged ground truth, and the corresponding rendered code artifacts from \textbf{VHarness (Ours)} and the \textbf{Baseline}. Red dashed lines highlight key layout boundaries and geometric contours. Compared with the baseline, VHarness faithfully preserves curved text wrapping in HTML, isometric contours and fine details in SVG, and organic branching geometry in 3D.}
    \label{fig:architecture}
  \end{figure*}%
}

\newcommand{\vhcase}{%
  \begin{figure*}[!t]
    \centering
    \includegraphics[width=0.85\textwidth]{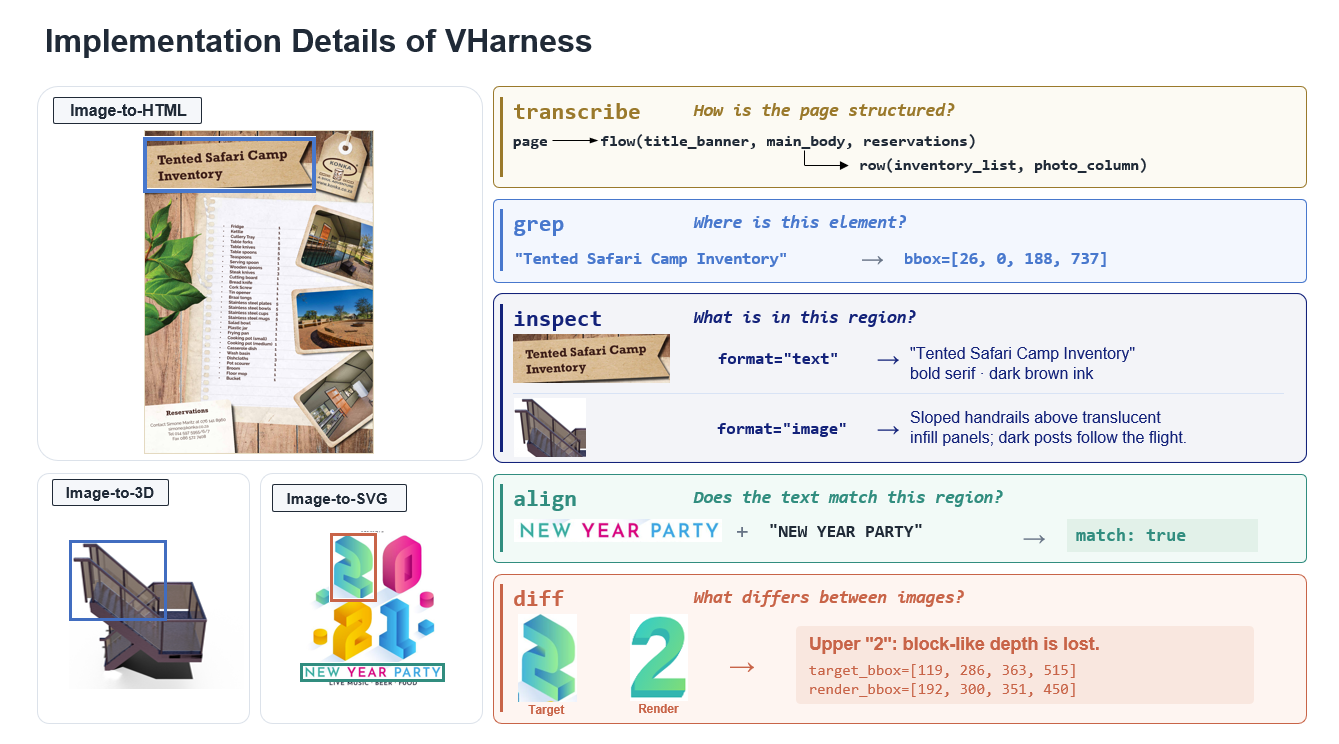}
    \caption{\textbf{Illustrative examples of the five atomic actions in VHarness across Image-to-HTML, Image-to-SVG, and Image-to-3D tasks.} Using representative visual inputs and highlighted regions on the left, the right panel illustrates the input queries and outputs of each callable action.}
    \label{fig:qualitative_cases}
  \end{figure*}%
}

\tcbset{vhpromptbox/.append style={top=3pt,bottom=3pt}}

\setcitestyle{numbers,square,comma}
\icmltitlerunning{Vision-as-a-Tool: Visual Harness for Multimodal Agentic Coding}

\author{%
  \parbox{\textwidth}{%
    \centering
    \normalsize
    \textbf{Haolin Yang}\textsuperscript{1,$\dagger$}\hspace{1.1em}
    \textbf{Wenshan Wu}\textsuperscript{2}\hspace{1.1em}
    \textbf{Jin Jiang}\textsuperscript{1}\hspace{1.1em}
    \textbf{Yu Cheng}\textsuperscript{3}\hspace{1.1em}
    \textbf{Yan Xia}\textsuperscript{2}\hspace{1.1em}
    \textbf{Ting Song}\textsuperscript{2}\\[2pt]
    \textbf{Yangyu Huang}\textsuperscript{2}\hspace{1.1em}
    \textbf{Zhiliang Peng}\textsuperscript{2}\hspace{1.1em}
    \textbf{Wenhui Wang}\textsuperscript{2}\hspace{1.1em}
    \textbf{Hangbo Bao}\textsuperscript{2}\hspace{1.1em}
    \textbf{Yuzhe Zhang}\textsuperscript{1}\\[2pt]
    \textbf{Wen-Da Wei}\textsuperscript{4}\hspace{1.1em}
    \textbf{Tengchao Lv}\textsuperscript{2}\hspace{1.1em}
    \textbf{Li Dong}\textsuperscript{2}\hspace{1.1em}
    \textbf{Lei Cui}\textsuperscript{2}\hspace{1.1em}
    \textbf{Furu Wei}\textsuperscript{2}\\[5pt]
    {\small \textbf{Project Page:}~\href{https://vision-as-a-tool.github.io/}{\textcolor{cvprblue}{\texttt{https://vision-as-a-tool.github.io/}}}}\\[5pt]
    \textsuperscript{1}Peking University\hspace{1.6em}
    \textsuperscript{2}Microsoft Research\hspace{1.6em}
    \textsuperscript{3}University of Edinburgh\hspace{1.6em}
    \textsuperscript{4}Nanjing University%
  }%
}

\hypersetup{
  pdfkeywords={Multimodal Large Language Models, Agentic Coding,
    Vision-to-Code Generation, Tool Use}
}

\renewcommand{\toptitlebar}{\hrule height1pt\vskip .10in}
\renewcommand{\bottomtitlebar}{\vskip .10in\hrule height1pt\vskip .12in}
\makeatletter
\newcommand{\vhprintauthors}{\par\begingroup\centering\@author\par\endgroup\vspace{4pt}}
\icml@noticeprintedtrue
\makeatother

\begin{document}

\twocolumn[
  \icmltitle{Vision-as-a-Tool: Visual Harness for Multimodal Agentic Coding}
  \vhprintauthors
  \begin{abstract}
Multimodal agentic coding requires robust visual reasoning to analyze observations and evaluate generated artifacts. Current frameworks largely rely on a \emph{Thinking-with-Images} paradigm, where agents depend on their intrinsic vision to execute visually grounded tasks. However, this architectural coupling introduces a fundamental bottleneck. Agent performance remains strictly bounded by internal perceptual limitations. Furthermore, text-only agents lack native perception. Without specialized tools, they cannot perform visual reasoning.
To overcome these constraints, we introduce the \textbf{Vision-as-a-Tool} (VaaT) paradigm. Prior visual tool-use merely manipulates images for an agent's intrinsic vision to read. In contrast, VaaT reframes general-purpose multimodal large language models (MLLMs) as callable external utilities. As a result, text-only agents can completely offload visual reasoning during agentic execution. Meanwhile, vision-capable agents can dynamically augment their native perception on demand.
We instantiate this paradigm in VHarness, a unified framework driven by five atomic actions. We comprehensively evaluate VHarness across Image-to-HTML, Image-to-SVG, and Image-to-3D generation tasks. Experimental results demonstrate that VHarness consistently improves visual fidelity and geometric accuracy across both text-only and multimodal coding agents. These findings highlight that modularizing visual reasoning into callable utilities provides a flexible and scalable foundation for future agentic coding systems.
\end{abstract}
  \vhfirstpageteaser
]
{\renewcommand{\thefootnote}{\fnsymbol{footnote}}%
 \footnotetext[2]{Work done during internship at Microsoft Research.}}
\clearpage

\section{Introduction}
\label{sec:introduction}

Multimodal agentic coding requires agents to interleave code generation with visual reasoning to analyze inputs and verify rendered artifacts. Current frameworks predominantly follow the \emph{Thinking-with-Images} paradigm~\citep{openai2025thinking,hu2024sketchpad,si2025design2code}, where coding agents rely on their own intrinsic vision to inspect images, whether directly or through basic image-manipulation operations such as cropping and zooming. However, tightly coupling visual perception with the coding agent introduces two fundamental limitations. First, for multimodal coding agents, output quality remains strictly bounded by their intrinsic perception. Even when recent studies design task-specific workflows~\citep{yuan2025designrepair,wu2025chat2svg,huang2024blenderalchemy} for iterative code refinement, they fail to break this intrinsic perceptual bound while sacrificing generalizability across diverse coding tasks. Second, text-only coding agents are entirely excluded from this paradigm due to the lack of native perception; while prior works attempt to equip them with fixed visual modules such as captioners and layout detectors~\citep{yang2023mmreact,suris2023vipergpt}, or perception pipelines tailored exclusively for visual question answering~\citep{zhang2025seeingeye,jia2025drp}, such designs cannot support the closed-loop refinement required by agentic coding.

To overcome these limitations, we introduce the \textbf{Vision-as-a-Tool (VaaT)} paradigm for multimodal agentic coding (Figure~\ref{fig:teaser}). Rather than relying on the agent's intrinsic vision or fixed task-specific perception designs, VaaT reframes an external MLLM as general-purpose visual utilities through a unified interface. Within this paradigm, the coding agent autonomously invokes these utilities during execution to obtain perceptual feedback that guides code generation and refinement. By decoupling visual perception from code generation, VaaT allows text-only coding agents to offload visual perception entirely to external visual utilities, while enabling multimodal coding agents to dynamically augment their intrinsic vision on demand.

To instantiate this paradigm, we develop \textbf{VHarness}, a unified framework organized around five intuitive atomic actions that expose visual utilities to the coding agent. Just as software engineering relies on general-purpose shell utilities such as \texttt{grep} for searching text and \texttt{diff} for comparing files, VHarness extends this philosophy to the visual domain: \texttt{transcribe} for holistic structure extraction, \texttt{grep} for visual element localization, \texttt{inspect} for fine-grained attribute querying, \texttt{align} for visual assertion checking, and \texttt{diff} for image-to-image comparison between target and rendered artifacts. This compact interface seamlessly adapts across diverse tasks without task-specific workflow engineering.

To rigorously and quantitatively evaluate the VaaT paradigm, we focus on vision-to-code generation tasks where rendered outputs can be directly verified against ground-truth visual inputs. Specifically, we evaluate VHarness across three diverse domains, spanning webpage reconstruction in Image-to-HTML, vector graphics generation in Image-to-SVG, and procedural modeling in Image-to-3D~\citep{gao2026codebench}. Our experiments verify the effectiveness of VHarness across two categories of coding agents. First, VHarness consistently improves visual fidelity and geometric accuracy across multimodal coding agents, including GPT-5.5~\citep{openai2026gpt55}, GPT-5.6-sol~\citep{openai2026gpt56}, Claude Opus 4.8~\citep{anthropic2026opus48}, and GLM-5.3-Flash~\citep{zai2026glm53flash}. Second, for text-only coding agents, VHarness enables GLM-5.2~\citep{zai2026glm52} to execute visually grounded tasks and achieve performance competitive with the native multimodal baseline GPT-5.5. Together, these results demonstrate that modularizing visual perception into callable utilities provides a flexible and scalable foundation for multimodal agentic coding.

The main contributions of this paper include:
\begin{itemize}
    \item \textbf{Vision-as-a-Tool Paradigm.} We propose the \textbf{Vision-as-a-Tool (VaaT)} paradigm for multimodal agentic coding, which reframes an external MLLM as general-purpose visual utilities, enabling coding agents to offload or augment visual perception on demand.
    \item \textbf{Unified VHarness Framework.} We develop \textbf{VHarness} around five intuitive atomic actions (\texttt{transcribe}, \texttt{grep}, \texttt{inspect}, \texttt{align}, and \texttt{diff}), providing a compact interface that adapts across diverse vision-to-code tasks under a unified action interface.
    \item \textbf{Cross-Task and Cross-Model Validation.} We validate VHarness across Image-to-HTML, Image-to-SVG, and Image-to-3D tasks, demonstrating consistent gains for both multimodal and text-only coding agents.
\end{itemize}
\section{Methodology}
\label{sec:methodology}

\subsection{Formulation of VaaT Paradigm}
\label{sec:formulation}

We formulate the process of multimodal agentic coding as follows. Given a visual input $x$ and a rendering environment $\mathcal{R}$, the coding agent is expected to iteratively generate verbal thoughts and code updates as part of the coding process and ultimately produce the final code. Let $\mathcal{P}_{\theta}$ represent a coding agent with parameters $\theta$, $\mathcal{U}_{\phi}$ denote callable external visual utilities with parameters $\phi$, and $z$, $c$, and $v$ denote the sequences of verbal thoughts, code updates, and rendered images, respectively, where $\hat{z}_i$, $\hat{c}_i$, and $\hat{v}_i = \mathcal{R}(\hat{c}_i)$ denote the corresponding elements at iteration $i$.

In multimodal agentic coding, at iteration $t$, let $V_t \subseteq \{x, \hat{v}_{1:t}\}$ denote the subset of selected images. Existing \emph{Thinking-with-Images} methods apply pixel-level image manipulations $\tau(\cdot)$ (e.g., cropping or zooming) to $V_t$ and feed the manipulated images back into the coding agent. \textbf{Vision-as-a-Tool (VaaT)} instead invokes the external visual utilities on demand with an atomic action $a_t$ on $V_t$ to obtain perceptual feedback, as shown in Figure~\ref{fig:teaser}. As defined in Equations~\ref{eq:twi} and~\ref{eq:vaat}, the code generation at iteration $t$ under the two paradigms is formulated as:
\begin{align}
    \hat{c}_{t+1} &\sim \mathcal{P}_{\theta}\big(c_{t+1} \mid x, \hat{z}_{1:t}, \hat{c}_{1:t}, \hat{v}_{1:t}, \tau(V_t)\big), \label{eq:twi} \\[3pt]
    \hat{c}_{t+1} &\sim \mathcal{P}_{\theta}\big(c_{t+1} \mid x, \hat{z}_{1:t}, \hat{c}_{1:t}, \hat{v}_{1:t}, \tau(V_t), \mathcal{U}_{\phi}(a_t, V_t)\big). \label{eq:vaat}
\end{align}

While \emph{Thinking-with-Images} relies on the coding agent's intrinsic vision to inspect the manipulated images, VaaT decouples visual perception from code generation by invoking the external visual utilities. Consequently, VaaT allows text-only coding agents that cannot accept image inputs $(x, \hat{v}_{1:t})$ to offload visual perception entirely to the external visual utilities, while enabling multimodal coding agents to dynamically augment their intrinsic vision on demand.

\subsection{Instantiating the VaaT Paradigm via VHarness}
\label{sec:action_space}

As shown in Figure~\ref{fig:qualitative_cases}, we present \textbf{VHarness} as a representative instantiation of the VaaT paradigm that implements external visual utilities using a general-purpose MLLM. Rather than designing task-specific perception pipelines, VHarness draws inspiration from traditional software engineering, where developers rely on general-purpose shell and debugging primitives to search patterns, inspect code blocks, check assertions, and compare file revisions. Extending this philosophy to the visual domain, VHarness exposes five intuitive atomic actions:
\begin{itemize}
    \item \texttt{transcribe}. For holistic structure extraction, \texttt{transcribe} takes the target image and generates a global layout description or initial structural draft to bootstrap code construction. 
    \item \texttt{grep}. For visual element localization, \texttt{grep} takes an image and a textual description, locates the corresponding visual elements, and returns their bounding boxes.
    \item \texttt{inspect}. For fine-grained visual observation, \texttt{inspect} takes an image or a localized region and returns detailed observations of local appearance, typography, or geometric structure.
    \item \texttt{align}. For visual assertion checking, given an image or a localized region together with a target text or specification, \texttt{align} returns a Boolean judgment indicating whether the visual content matches the specification.
    \item \texttt{diff}. For image-to-image comparison, given the target image and the rendered image, \texttt{diff} identifies visual discrepancies between them and localizes mismatched regions to guide subsequent refinement.
\end{itemize}

\vharchitecture

\subsection{On-Demand Agentic Execution}
\label{sec:execution_loop}

Rather than prescribing a fixed workflow, VHarness exposes the five atomic actions as callable tools (Figure~\ref{fig:qualitative_cases}), allowing the coding agent to autonomously coordinate code updates, intrinsic vision, and external visual utilities throughout the coding process. During execution, both text-only and multimodal coding agents can invoke \texttt{grep} and \texttt{inspect} on demand to resolve regional placement or fine-grained visual details. Once the code is rendered into an image, the agent flexibly interleaves \texttt{align} and \texttt{diff} with further \texttt{grep} and \texttt{inspect} calls to diagnose discrepancies and plan subsequent refinements. Through this dynamic interplay, the agent continuously translates perceptual feedback into targeted code updates.

In practice, VHarness exposes the five atomic actions as callable tools through the Model Context Protocol (MCP). Crucially, neither the baseline nor VHarness enforces a minimum iteration count or fixed token budget; instead, both terminate once the rendered artifact is judged visually complete. Whereas a baseline agent relies solely on its intrinsic vision to decide when to stop, an agent with VHarness relies on external visual utilities to verify completion.


\section{Experiments and Analysis}
\label{sec:experiments}

\begin{table*}[!t]
    \centering
    \caption{Quantitative comparison of final rendered artifacts with (\texttt{w/}) and without (\texttt{w/o}) VHarness across Image-to-HTML, Image-to-SVG, and Image-to-3D (100 cases per task). $^*$Due to inference budget constraints, the text-only coding agent GLM-5.2 is evaluated on a random subset of 50 cases per task. Arrows indicate the preferred direction for each metric, and bold marks the best result in each column.}
    \label{tab:main_results}
    \newcommand{\vhbest}[1]{\textbf{#1}}
    \newcommand{\mc}[1]{\makebox[0.08666\textwidth][c]{#1}}
    \newcommand{\gc}[1]{\makebox[0.13\textwidth][c]{#1}}
    \newcommand{\modcol}[1]{\makebox[0.15\textwidth][l]{#1}}
    \newcommand{\vhcol}[1]{\makebox[0.07\textwidth][c]{#1}}
    \footnotesize
    \setlength{\tabcolsep}{0pt}
    \renewcommand{\arraystretch}{1.16}
    \begin{tabular*}{\textwidth}{@{}llccccccccc@{}}
        \toprule
        \multicolumn{11}{@{}l}{\textbf{(a) Image-to-HTML}} \\
        \modcol{Model} & \vhcol{VHarness} & \multicolumn{3}{c}{2D Perceptual} & \multicolumn{6}{c}{2D Structural} \\
        \cmidrule(lr){3-5}\cmidrule(l){6-11}
        & & \mc{SigLIP2 $\uparrow$} & \mc{DINOv2 $\uparrow$} & \mc{LPIPS $\downarrow$} & \mc{SSIM l. $\uparrow$} & \mc{SSIM g. $\uparrow$} & \mc{Px MSE $\downarrow$} & \mc{Edge P $\uparrow$} & \mc{Edge R $\uparrow$} & \mc{Edge F1 $\uparrow$} \\
        \midrule
        \modcol{GPT-5.5} & \vhcol{w/o} & \mc{0.5773} & \mc{0.8881} & \mc{0.2277} & \mc{0.6789} & \mc{0.6950} & \mc{0.0500} & \mc{0.5464} & \mc{0.5335} & \mc{0.5389} \\
         & \vhcol{w/} & \mc{0.6319} & \mc{0.9046} & \mc{0.1820} & \mc{0.6930} & \mc{0.7201} & \mc{0.0461} & \mc{0.5867} & \mc{0.5792} & \mc{0.5820} \\
        \addlinespace[1.6pt]
        \modcol{GPT-5.6-sol} & \vhcol{w/o} & \mc{0.6527} & \mc{0.9055} & \mc{0.1635} & \mc{0.7197} & \mc{0.7496} & \mc{0.0414} & \mc{0.6625} & \mc{0.6433} & \mc{0.6517} \\
         & \vhcol{w/} & \mc{0.6859} & \mc{0.9131} & \mc{0.1414} & \mc{0.7385} & \mc{0.7663} & \mc{0.0385} & \mc{0.7009} & \mc{0.6883} & \mc{0.6936} \\
        \addlinespace[1.6pt]
        \modcol{Claude Opus 4.8} & \vhcol{w/o} & \mc{0.3908} & \mc{0.7167} & \mc{0.4596} & \mc{0.5494} & \mc{0.4539} & \mc{0.0896} & \mc{0.3395} & \mc{0.3152} & \mc{0.3257} \\
         & \vhcol{w/} & \mc{0.5077} & \mc{0.8715} & \mc{0.2974} & \mc{0.6084} & \mc{0.6211} & \mc{0.0615} & \mc{0.4399} & \mc{0.4251} & \mc{0.4316} \\
        \addlinespace[1.6pt]
        \modcol{GLM-5.2$^*$} & \vhcol{w/} & \mc{0.6044} & \mc{0.8949} & \mc{0.2028} & \mc{0.6715} & \mc{0.7242} & \mc{0.0477} & \mc{0.6107} & \mc{0.5736} & \mc{0.5907} \\
        \addlinespace[1.6pt]
        \modcol{GLM-5.3-Flash} & \vhcol{w/o} & \mc{0.6743} & \mc{0.9019} & \mc{0.1486} & \mc{0.7411} & \mc{0.7696} & \mc{0.0387} & \mc{0.7186} & \mc{0.7020} & \mc{0.7089} \\
         & \vhcol{w/} & \mc{\vhbest{0.7553}} & \mc{\vhbest{0.9216}} & \mc{\vhbest{0.0937}} & \mc{\vhbest{0.7942}} & \mc{\vhbest{0.8196}} & \mc{\vhbest{0.0297}} & \mc{\vhbest{0.8135}} & \mc{\vhbest{0.8003}} & \mc{\vhbest{0.8059}} \\
        \bottomrule
    \end{tabular*}

    \par\vspace{0.6em}
    \begin{tabular*}{\textwidth}{@{}llccccccccc@{}}
        \toprule
        \multicolumn{11}{@{}l}{\textbf{(b) Image-to-SVG}} \\
        \modcol{Model} & \vhcol{VHarness} & \multicolumn{3}{c}{2D Perceptual} & \multicolumn{6}{c}{2D Structural} \\
        \cmidrule(lr){3-5}\cmidrule(l){6-11}
        & & \mc{SigLIP2 $\uparrow$} & \mc{DINOv2 $\uparrow$} & \mc{LPIPS $\downarrow$} & \mc{SSIM l. $\uparrow$} & \mc{SSIM g. $\uparrow$} & \mc{Px MSE $\downarrow$} & \mc{Edge P $\uparrow$} & \mc{Edge R $\uparrow$} & \mc{Edge F1 $\uparrow$} \\
        \midrule
        \modcol{GPT-5.5} & \vhcol{w/o} & \mc{0.6265} & \mc{0.8524} & \mc{0.1537} & \mc{0.8146} & \mc{0.8037} & \mc{0.0159} & \mc{0.6239} & \mc{0.6068} & \mc{0.6141} \\
         & \vhcol{w/} & \mc{0.6537} & \mc{0.8793} & \mc{0.1266} & \mc{0.8281} & \mc{0.8375} & \mc{0.0132} & \mc{0.6561} & \mc{0.6428} & \mc{0.6478} \\
        \addlinespace[1.6pt]
        \modcol{GPT-5.6-sol} & \vhcol{w/o} & \mc{0.6703} & \mc{0.8884} & \mc{0.1087} & \mc{0.8421} & \mc{0.8614} & \mc{0.0116} & \mc{0.7052} & \mc{0.6938} & \mc{0.6985} \\
         & \vhcol{w/} & \mc{0.6830} & \mc{0.9020} & \mc{0.1001} & \mc{0.8502} & \mc{0.8652} & \mc{0.0108} & \mc{0.7055} & \mc{0.7001} & \mc{0.7020} \\
        \addlinespace[1.6pt]
        \modcol{Claude Opus 4.8} & \vhcol{w/o} & \mc{0.4774} & \mc{0.7220} & \mc{0.3759} & \mc{0.7100} & \mc{0.4951} & \mc{0.0443} & \mc{0.3286} & \mc{0.3088} & \mc{0.3165} \\
         & \vhcol{w/} & \mc{0.5502} & \mc{0.7871} & \mc{0.2605} & \mc{0.7551} & \mc{0.6729} & \mc{0.0281} & \mc{0.4510} & \mc{0.4257} & \mc{0.4360} \\
        \addlinespace[1.6pt]
        \modcol{GLM-5.2$^*$} & \vhcol{w/} & \mc{0.6507} & \mc{0.8656} & \mc{0.1219} & \mc{0.8471} & \mc{0.8756} & \mc{0.0094} & \mc{0.7383} & \mc{0.6665} & \mc{0.6950} \\
        \addlinespace[1.6pt]
        \modcol{GLM-5.3-Flash} & \vhcol{w/o} & \mc{0.6887} & \mc{0.8974} & \mc{0.0943} & \mc{0.8597} & \mc{0.8880} & \mc{0.0085} & \mc{0.7292} & \mc{0.7245} & \mc{0.7262} \\
         & \vhcol{w/} & \mc{\vhbest{0.7095}} & \mc{\vhbest{0.9207}} & \mc{\vhbest{0.0778}} & \mc{\vhbest{0.8742}} & \mc{\vhbest{0.9109}} & \mc{\vhbest{0.0069}} & \mc{\vhbest{0.7598}} & \mc{\vhbest{0.7765}} & \mc{\vhbest{0.7671}} \\
        \bottomrule
    \end{tabular*}

    \par\vspace{0.6em}
    \begin{tabular*}{\textwidth}{@{}llcccccccc@{}}
        \toprule
        \multicolumn{10}{@{}l}{\textbf{(c) Image-to-3D}} \\
        \modcol{Model} & \vhcol{VHarness} & \multicolumn{3}{c}{2D Perceptual} & \multicolumn{3}{c}{2D Structural} & \multicolumn{2}{c}{3D Geometric} \\
        \cmidrule(lr){3-5}\cmidrule(lr){6-8}\cmidrule(l){9-10}
        & & \mc{SigLIP2 $\uparrow$} & \mc{DINOv2 $\uparrow$} & \mc{LPIPS $\downarrow$} & \mc{SSIM l. $\uparrow$} & \mc{SSIM g. $\uparrow$} & \mc{Px MSE $\downarrow$} & \gc{Uni3D $\uparrow$} & \gc{Chamfer $\downarrow$} \\
        \midrule
        \modcol{GPT-5.5} & \vhcol{w/o} & \mc{0.6101} & \mc{0.6622} & \mc{0.1485} & \mc{0.8901} & \mc{0.7157} & \mc{0.0242} & \gc{0.7575} & \gc{0.0094} \\
         & \vhcol{w/} & \mc{0.6822} & \mc{0.7496} & \mc{0.0875} & \mc{0.9190} & \mc{0.8541} & \mc{0.0092} & \gc{0.7933} & \gc{0.0058} \\
        \addlinespace[1.6pt]
        \modcol{GPT-5.6-sol} & \vhcol{w/o} & \mc{0.6400} & \mc{0.7046} & \mc{0.1211} & \mc{0.9043} & \mc{0.7515} & \mc{0.0169} & \gc{0.7738} & \gc{0.0157} \\
         & \vhcol{w/} & \mc{0.7212} & \mc{0.8122} & \mc{0.0634} & \mc{0.9296} & \mc{0.9023} & \mc{0.0055} & \gc{0.8496} & \gc{\vhbest{0.0035}} \\
        \addlinespace[1.6pt]
        \modcol{Claude Opus 4.8} & \vhcol{w/o} & \mc{0.5194} & \mc{0.5342} & \mc{0.2353} & \mc{0.8738} & \mc{0.4586} & \mc{0.0384} & \gc{0.7129} & \gc{0.0339} \\
         & \vhcol{w/} & \mc{0.6344} & \mc{0.7017} & \mc{0.1269} & \mc{0.9048} & \mc{0.7502} & \mc{0.0149} & \gc{0.7863} & \gc{0.0086} \\
        \addlinespace[1.6pt]
        \modcol{GLM-5.2$^*$} & \vhcol{w/} & \mc{0.6400} & \mc{0.6623} & \mc{0.1250} & \mc{0.9076} & \mc{0.7470} & \mc{0.0162} & \gc{0.7548} & \gc{0.0092} \\
        \addlinespace[1.6pt]
        \modcol{GLM-5.3-Flash} & \vhcol{w/o} & \mc{0.6937} & \mc{0.7593} & \mc{0.0838} & \mc{0.9229} & \mc{0.8612} & \mc{0.0082} & \gc{0.8154} & \gc{0.0056} \\
         & \vhcol{w/} & \mc{\vhbest{0.7336}} & \mc{\vhbest{0.8190}} & \mc{\vhbest{0.0567}} & \mc{\vhbest{0.9367}} & \mc{\vhbest{0.9262}} & \mc{\vhbest{0.0045}} & \gc{\vhbest{0.8545}} & \gc{\vhbest{0.0035}} \\
        \bottomrule
    \end{tabular*}
\end{table*}

\subsection{Vision-to-Code Tasks}
\label{sec:tasks}

To rigorously and quantitatively evaluate the VaaT paradigm in multimodal agentic coding, we focus on vision-to-code generation tasks, where rendered outputs can be directly verified against ground-truth visual inputs. While prior benchmarks such as Design2Code~\citep{si2025design2code} have been widely used for visual coding evaluation, their test cases predominantly consist of standard grid-based web templates or simple graphic icons, on which modern MLLMs and even fine-tuned lightweight models can already solve with high accuracy. To provide sufficient difficulty to demonstrate the effectiveness of VHarness, we benchmark across three tasks spanning complex Image-to-HTML, challenging Image-to-SVG, and Image-to-3D, each comprising 100 complex test cases (see Appendix~\ref{app:implementation} for data curation details and cases).

\paragraph{Image-to-HTML.}
We curate 100 complex webpage designs spanning diverse layout styles from publicly available online sources, encompassing not only complex webpages but also editorial magazines, promotional posters, and presentation slides with highly irregular structures. For each case, the coding agent receives a full-page target screenshot accompanied by pre-extracted visual assets, such as icons and illustrations, and generates a self-contained HTML/CSS document rendered by a headless browser. All models and experimental configurations receive the same assets without spatial coordinates or layout metadata. This task primarily evaluates the agent's ability to recover text flow and document hierarchy, requiring it to jointly reconstruct structural layout, typography, asset placement, and complex text-wrapping behaviors solely from visual observation. Representative benchmark cases and experimental setup details are provided in Appendices~\ref{app:implementation} and~\ref{app:experimental_setup}.

\paragraph{Image-to-SVG.}
We collect 100 graphic designs from publicly available online sources, focusing on examples with intricate geometry and fine-grained detail compositions. Given a raster target image without any external visual assets, the coding agent must generate standalone Scalable Vector Graphics (SVG) code, which the rendering environment rasterizes into a rendered image. Unlike webpage reconstruction, Image-to-SVG requires the agent to synthesize continuous shapes, B\'ezier paths, gradients, and multi-layered detail compositions purely through vector primitives. Representative benchmark cases and experimental setup details are provided in Appendices~\ref{app:implementation} and~\ref{app:experimental_setup}.

\paragraph{Image-to-3D.}
We randomly sample 100 cases from 3DCodeBench~\citep{gao2026codebench}, where the target visual input consists of four canonical views of a 3D object. The coding agent generates an executable Blender Python script to procedurally construct the 3D mesh, which the rendering environment exports in GLB format and renders from matching camera viewpoints to produce the rendered views. This task extends visual coding from 2D planes to 3D space, requiring the agent to infer volumetric proportions, part connectivity, and multi-view geometric consistency from 2D projections. Representative benchmark cases and experimental setup details are provided in Appendices~\ref{app:implementation} and~\ref{app:experimental_setup}.

\subsection{Experimental Setup}
\label{sec:experimental_setup}

\paragraph{Models and Baselines.}
To ensure a controlled comparison, both the baseline (\texttt{w/o VHarness}) and our method (\texttt{w/ VHarness}) are built upon the standard Claude Code agent harness. Under the baseline setting, vision-capable agents follow the \emph{Thinking-with-Images} paradigm within the default harness environment, where they can use pre-installed Python image-processing libraries or install additional packages on demand to assist their native visual inspection. Building directly upon this baseline harness, VHarness registers the five atomic visual actions as callable MCP tools, using Gemini-3.8-Flash to power the external visual utilities. Within this setup, we evaluate two categories of coding agents, encompassing multimodal coding agents (GPT-5.5~\citep{openai2026gpt55}, GPT-5.6-sol~\citep{openai2026gpt56}, Claude Opus 4.8~\citep{anthropic2026opus48}, and GLM-5.3-Flash~\citep{zai2026glm53flash}) and the text-only coding agent GLM-5.2~\citep{zai2026glm52}. Because GLM-5.2 lacks native vision to process image inputs directly, it cannot run under the baseline settings and is evaluated exclusively with VHarness.

\paragraph{Evaluation Metrics.}
We evaluate the final outputs using three groups of offline metrics that measure 2D perceptual fidelity, 2D structural fidelity, and 3D geometric accuracy. The 2D perceptual group assesses semantic and feature alignment through SigLIP2~\citep{tschannen2025siglip2} and DINOv2~\citep{oquab2023dinov2} similarities as well as LPIPS perceptual distance~\citep{zhang2018lpips}. The 2D structural group measures spatial layout, pixel accuracy, and contour alignment using local and global SSIM~\citep{wang2004ssim}, Pixel MSE, and Canny edge metrics. The 3D geometric group evaluates mesh reconstruction quality using Uni3D similarity~\citep{zhou2023uni3d} and normalized squared Chamfer distance~\citep{fan2017pointset}, following the official evaluation protocol of 3DCodeBench~\citep{gao2026codebench}. As summarized in Table~\ref{tab:main_results}, we evaluate Image-to-HTML and Image-to-SVG with the 2D perceptual and structural metrics, while additionally incorporating the 3D geometric metrics for Image-to-3D (see Appendix B for hyperparameter configurations and metric definitions).

\vhcase

\subsection{Main Results and Qualitative Analysis}
\label{sec:main_results}

\paragraph{Consistent Gains Across Multimodal Coding Agents.}
As shown in Table~\ref{tab:main_results}, VHarness consistently improves every evaluation metric across all four multimodal coding agents on all three tasks. Beyond uniform within-model gains, VHarness substantially narrows performance gaps across model tiers on 2D tasks and even reverses them on Image-to-3D. Specifically, GPT-5.5 with VHarness surpass the stronger GPT-5.6-sol baseline across all metrics on Image-to-3D. Moreover, even on GLM-5.3-Flash, which already establishes the strongest baseline across all three tasks, VHarness still delivers substantial further improvements and achieves the best overall performance in every column of Table~\ref{tab:main_results}. These results demonstrate that augmenting intrinsic vision with external visual utilities effectively breaks through internal perceptual bottlenecks across multimodal coding agents of varying capabilities.

\paragraph{Empowering Text-Only Coding Agents to Compete with Multimodal Baselines.}
Equipped with VHarness, the text-only coding agent GLM-5.2 successfully executes all three vision-to-code tasks and achieves highly competitive performance. Specifically, GLM-5.2 with VHarness outperforms the multimodal Claude Opus 4.8 baseline across all three tasks. Moreover, GLM-5.2 with VHarness exceeds the GPT-5.5 baseline on nearly all HTML and SVG metrics, and remains competitive with the stronger GPT-5.6-sol baseline, outperforming it on multiple 2D structural metrics. This finding confirms that intrinsic vision is not a prerequisite for high-fidelity visual coding, as offloading visual perception to external visual utilities enables a strong text-only coding agent to match or exceed multimodal baselines.

\paragraph{Qualitative Analysis.}
Figure~\ref{fig:architecture} presents qualitative comparisons between the baseline and VHarness across all three tasks. In Image-to-HTML, the baseline fails to capture the circular text-wrapping layout and collapses into a rigid vertical column, whereas VHarness accurately reconstructs the curved text boundary around the central illustration. In Image-to-SVG, the baseline omits fine structural components and distorts the left and lower-right platform contours, while VHarness preserves both global geometry and local details. In Image-to-3D, the baseline degenerates into a simplistic cylinder with radial spikes, whereas VHarness faithfully reproduces the organic trunk curvature and branching structure of the target tree. Across all three domains, VHarness aligns substantially closer with the ground truth in both layout and geometry.

\begin{figure}[!t]
    \centering
    \includegraphics[width=0.85\columnwidth]
        {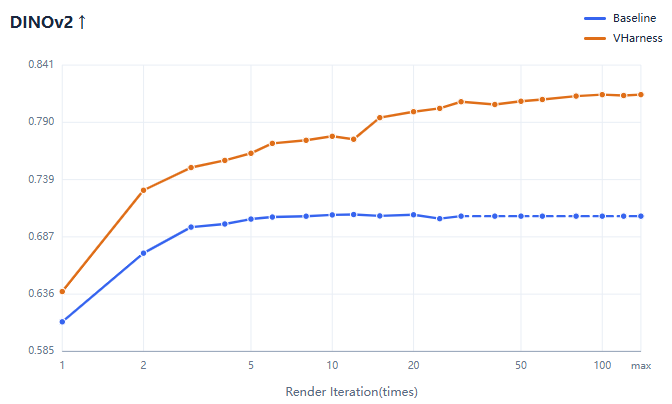}
    \caption{\textbf{Refinement trajectories of GPT-5.6-sol with DINOv2
    on Image-to-3D.} 
    Dashed segments extend the baseline's final scores
    beyond termination.}
    \label{fig:main_refinement_dynamics}
\end{figure}

To analyze how VHarness improves code quality across render cycles, Figure~\ref{fig:main_refinement_dynamics} shows that VHarness achieves higher DINOv2 scores at matched early render iterations and continues improving after the baseline terminates. Appendix~\ref{app:cost_and_cases} provides
additional models, metrics with analysis.

\subsection{Ablation Study}
\label{sec:ablation_study}

\begin{table}[!t]
    \centering
    \caption{Ablation study with GPT-6-astra on the Image-to-SVG benchmark. Bold marks the better result in each row.}
    \label{tab:astra_same_model}
    \small
    \setlength{\tabcolsep}{6pt}
    \renewcommand{\arraystretch}{1.12}
    \begin{tabular*}{\columnwidth}{@{\extracolsep{\fill}}lrr@{}}
        \toprule
        Metric & \texttt{w/o VHarness} & \texttt{w/ VHarness} \\
        \midrule
        SigLIP2 $\uparrow$   & 0.7368 & \textbf{0.7529} \\
        DINOv2 $\uparrow$    & 0.9428 & \textbf{0.9486} \\
        LPIPS $\downarrow$   & 0.0572 & \textbf{0.0473} \\
        \midrule
        SSIM l. $\uparrow$   & 0.8903 & \textbf{0.8987} \\
        SSIM g. $\uparrow$   & 0.9229 & \textbf{0.9356} \\
        Px MSE $\downarrow$  & 0.0061 & \textbf{0.0051} \\
        Edge P $\uparrow$   & 0.8337 & \textbf{0.8556} \\
        Edge R $\uparrow$  & 0.8299 & \textbf{0.8550} \\
        Edge F1 $\uparrow$   & 0.8310 & \textbf{0.8546} \\
        \bottomrule
    \end{tabular*}

    \vspace{10pt}

    \caption{Ablation study on replacing the five atomic actions with a single free-form MLLM interface (\texttt{Freestyle}) using GPT-5.6-sol on the Image-to-3D benchmark. Bold marks the best result in each row.}
    \label{tab:freestyle_ablation}
    \small
    \setlength{\tabcolsep}{0pt}
    \renewcommand{\arraystretch}{1.12}
    \begin{tabular*}{\columnwidth}{@{\extracolsep{\fill}}lrrr@{}}
        \toprule
        Metric & \texttt{w/o VHarness} & \texttt{Freestyle} & \texttt{w/ VHarness} \\
        \midrule
        SigLIP2 $\uparrow$   & 0.6400 & 0.6677 & \textbf{0.7212} \\
        DINOv2 $\uparrow$    & 0.7046 & 0.7481 & \textbf{0.8122} \\
        LPIPS $\downarrow$   & 0.1211 & 0.1028 & \textbf{0.0634} \\
        \midrule
        SSIM l. $\uparrow$   & 0.9043 & 0.9120 & \textbf{0.9296} \\
        SSIM g. $\uparrow$   & 0.7515 & 0.8150 & \textbf{0.9023} \\
        Px MSE $\downarrow$  & 0.0169 & 0.0111 & \textbf{0.0055} \\
        \midrule
        Uni3D $\uparrow$     & 0.7738 & 0.8255 & \textbf{0.8496} \\
        Chamfer $\downarrow$ & 0.0157 & 0.0070 & \textbf{0.0035} \\
        \bottomrule
    \end{tabular*}
\end{table}

While Table~\ref{tab:main_results} demonstrates consistent gains across diverse coding agents, a fundamental question remains: \textbf{whether the Vision-as-a-Tool (VaaT) paradigm stays effective as a model's intrinsic vision becomes stronger}. To investigate this question, we evaluate a stronger frontier multimodal model, GPT-6-astra, on our 100 cases Image-to-SVG benchmark. Under the \texttt{w/o VHarness} baseline, GPT-6-astra follows the \emph{Thinking-with-Images} paradigm using its own intrinsic vision, whereas under \texttt{w/ VHarness}, it serves as both the coding agent and the external visual utilities.

As shown in Table~\ref{tab:astra_same_model}, the GPT-6-astra baseline (\texttt{w/o VHarness}) already surpasses all models and configurations in Table~\ref{tab:main_results}(b) across every metric, reflecting its stronger intrinsic vision. Even at this frontier perceptual level, equipping GPT-6-astra with VHarness still yields consistent improvements across all perceptual and structural metrics. These results confirm that the VaaT paradigm remains effective as a model's intrinsic vision continues to strengthen.


Since VHarness instantiates the VaaT paradigm through five atomic actions, a natural design question is \textbf{whether atomic actions in VHarness can be simplified into a single unified MLLM interface}. To investigate this, we evaluate GPT-5.6-sol on the 100-case Image-to-3D benchmark under a \texttt{Freestyle} setting. Specifically, we replace the five atomic actions with a single general-purpose MCP tool backed by the same Gemini-3.8-Flash model and provide usage examples in the skill instructions (see Appendix~\ref{app:prompts_skills}), allowing the coding agent to freely query the external MLLM on any target or rendered views using self-formulated natural language prompts.

Inspecting the execution trajectories reveals that, guided by the usage examples, the coding agent under \texttt{Freestyle} spontaneously attempts queries analogous to our atomic actions, such as comparing multi-view discrepancies (\texttt{diff}) or examining local part geometry (\texttt{inspect}). However, without the five atomic actions of VHarness, the coding agent under \texttt{Freestyle} cannot invoke the external MLLM with accuracy and consistency just like it uses standard \texttt{bash} utilities. Consequently, as shown in Table~\ref{tab:freestyle_ablation}, while access to the external MLLM enables \texttt{Freestyle} to consistently outperform the \texttt{w/o VHarness} baseline across all 2D and 3D metrics, it still falls substantially short of \texttt{w/ VHarness} (e.g., $0.6677$ vs.\ $0.7212$ in SigLIP2 and $0.0070$ vs.\ $0.0035$ in Chamfer distance). These findings confirm that while offloading visual perception to external visual utilities is beneficial, encapsulating it into well-defined atomic actions is essential for accurate and effective iterative code refinement.

\section{Related Work}
\label{sec:relatedwork}

\subsection{Multimodal Agents and Visual Tool Use}
Current multimodal coding agents predominantly follow the \emph{Thinking-with-Images} paradigm~\citep{openai2025thinking}, where models rely on their own intrinsic vision to inspect visual inputs and intermediate states during execution. Following this paradigm, Visual Sketchpad~\citep{hu2024sketchpad} equips MLLMs with drawing and image-manipulation operations such as cropping and zooming, returning the manipulated images to the model's own visual encoder for re-observation. Similarly, Design2Code~\citep{si2025design2code} prompts MLLMs to directly inspect rendered webpage screenshots and self-revise the generated HTML/CSS code. Building upon this paradigm, recent studies design task-specific workflows to guide iterative code refinement across individual domains. Specifically, DesignRepair~\citep{yuan2025designrepair} couples rendered page analysis with design-guideline retrieval to repair frontend UI code; Chat2SVG~\citep{wu2025chat2svg} combines MLLM-based visual checks with diffusion-guided optimization to refine vector graphics templates; and BlenderAlchemy~\citep{huang2024blenderalchemy} iteratively edits Blender Python scripts by evaluating rendered 3D scenes with vision-language models. However, because visual interpretation in these approaches ultimately depends on the agent's intrinsic perception, their output quality remains strictly bounded by internal perceptual limits, while such rigid, task-specific workflows fail to generalize across diverse coding tasks.

Meanwhile, text-only agents must rely on external modules to process visual inputs. Early frameworks coordinate fixed, specialized vision models such as image captioners, OCR engines, and layout or object detectors: MM-REACT~\citep{yang2023mmreact}, Visual ChatGPT~\citep{wu2023visualchatgpt}, and HuggingGPT~\citep{shen2023hugginggpt} invoke these domain experts through natural language orchestration, whereas VISPROG~\citep{gupta2023visprog} and ViperGPT~\citep{suris2023vipergpt} compose predefined vision APIs by generating executable Python programs. More recently, VQA-oriented decoupled pipelines such as SeeingEye~\citep{zhang2025seeingeye} and DRP~\citep{jia2025drp} pair a text-only reasoner with a multimodal translator or observer to distill intermediate descriptions or query visual details during multi-step reasoning. Nevertheless, fixed specialist modules remain restricted to narrow perceptual primitives, while VQA-oriented pipelines are designed exclusively for static image comprehension and cannot support the iterative artifact verification and closed-loop code refinement required by agentic coding. To overcome these limitations, our \textbf{Vision-as-a-Tool (VaaT)} paradigm reframes an external MLLM as general-purpose callable visual utilities, enabling text-only coding agents to offload visual perception entirely while allowing multimodal coding agents to dynamically augment their intrinsic vision on demand across diverse coding tasks.

\subsection{Agentic Coding and Vision-to-Code Generation}
Multimodal agentic coding requires agents to interleave code generation with visual reasoning to analyze inputs and verify rendered artifacts, with vision-to-code generation serving as a quantitatively verifiable testbed. Existing studies have explored this vision-to-code problem across diverse artifact modalities. In 2D webpage engineering, early work such as pix2code~\citep{beltramelli2017pix2code} and Design2Code~\citep{si2025design2code} established benchmarks for translating screenshots into HTML and CSS, followed by methods incorporating rendered visual feedback for iterative code refinement~\citep{chen2025rrvf,li2026relook,deng2026visrefiner}. Moving from structured DOM layouts to continuous vector primitives, StarVector~\citep{rodriguez2025starvector} synthesizes Scalable Vector Graphics (SVG) code directly from raster images, while in 3D procedural modeling, CAD-Coder~\citep{doris2025cadcoder} and 3DCodeBench~\citep{gao2026codebench} generate modeling scripts from multi-view image references. Despite this progress, prior methods remain confined to individual artifact modalities with task-specific designs. Moreover, existing 2D benchmarks such as Design2Code and standard SVG datasets feature relatively simple layouts and icons that fall short of real-world complexity, on which both proprietary MLLMs and lightweight open-source models already achieve near-ceiling performance. To better demonstrate the effectiveness of VHarness across HTML, SVG, and 3D tasks, we curate substantially more challenging and realistic test cases spanning diverse real-world visual scenarios (see Appendix~\ref{app:implementation} for dataset details and representative cases).

\section{Conclusion and Limitations}
\label{sec:conclusion}

\paragraph{Conclusion.}
We presented \textbf{Vision-as-a-Tool (VaaT)}, a paradigm for multimodal agentic coding that reframes an external MLLM as general-purpose callable visual utilities. Drawing inspiration from traditional software engineering, where developers rely on shell and debugging primitives to search, inspect, assert, and compare files, we instantiated this paradigm in \textbf{VHarness} by extending this philosophy to the visual domain via five intuitive atomic actions. Across Image-to-HTML, Image-to-SVG, and Image-to-3D tasks, VHarness consistently improves the visual and geometric fidelity of multimodal coding agents, while enabling text-only coding agents to execute vision-to-code tasks and achieve performance competitive with native multimodal baselines. Together, these results demonstrate that decoupling visual perception into on-demand callable utilities provides a general and scalable foundation for multimodal agentic coding.

\paragraph{Limitations.}
While effective across diverse vision-to-code tasks, our study primarily evaluates static targets (single-page interfaces, standalone vector graphics, and individual 3D objects) rather than dynamic multi-state interactions. Extending VHarness to interactive applications and temporal 3D scenes remains an important future direction.

{
  \small
  \bibliographystyle{ieeenat_fullname}
  \bibliography{references}
}

\clearpage
\appendix
\raggedbottom
\section{Implementation Details of VHarness}
\label{app:implementation}

\subsection{Benchmark Curation and Execution Environments}
\label{app:environments}
\label{app:data_curation}

\begin{figure*}[!t]
    \centering
    \includegraphics[width=0.97\textwidth]{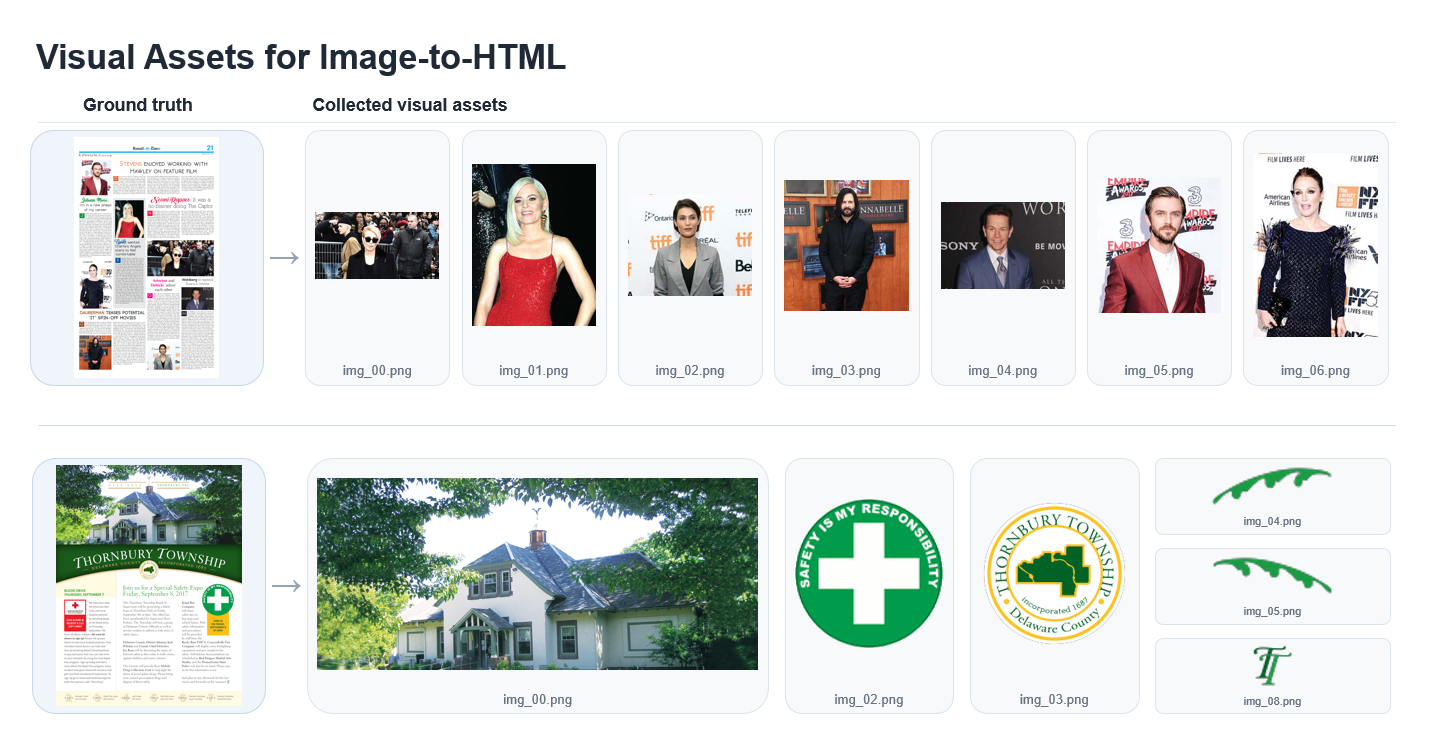}\par
    \vspace{6pt}
    \includegraphics[width=0.97\textwidth]{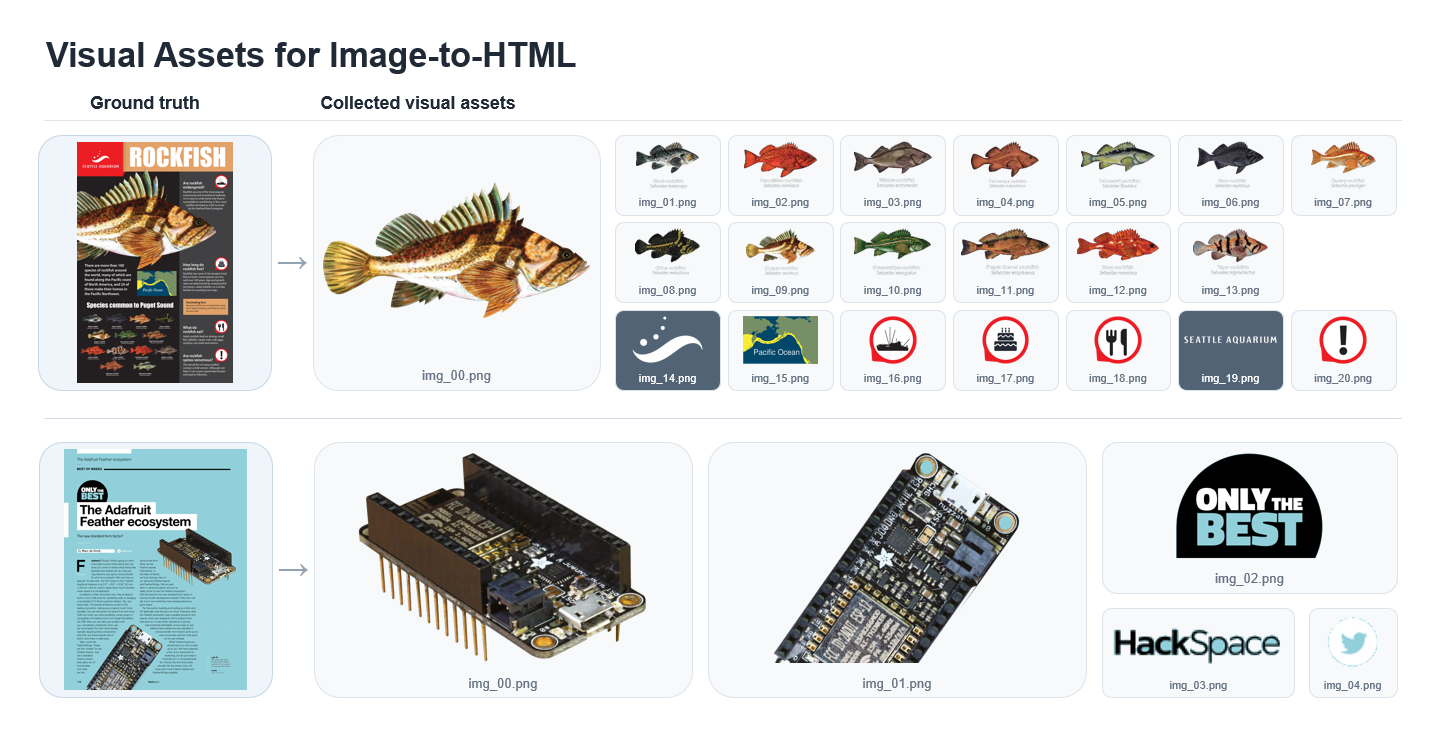}
    \caption{\textbf{Examples of workspace inputs provided to the coding agent in Image-to-HTML.} For each test case, the workspace supplies the target webpage screenshot (\emph{Visual Input}, \texttt{target.png}, left), the pre-extracted graphic files stored under \texttt{./assets/} (\emph{Visual Assets}, middle), and the JSON manifest (\texttt{provided\_assets.json}, right) specifying each asset's filename, relative path, and native pixel dimensions without spatial coordinates or layout metadata.}
    \label{fig:appendix_visual_assets}
\end{figure*}

\begin{figure*}[!t]
    \centering
    \includegraphics[width=0.94\textwidth]{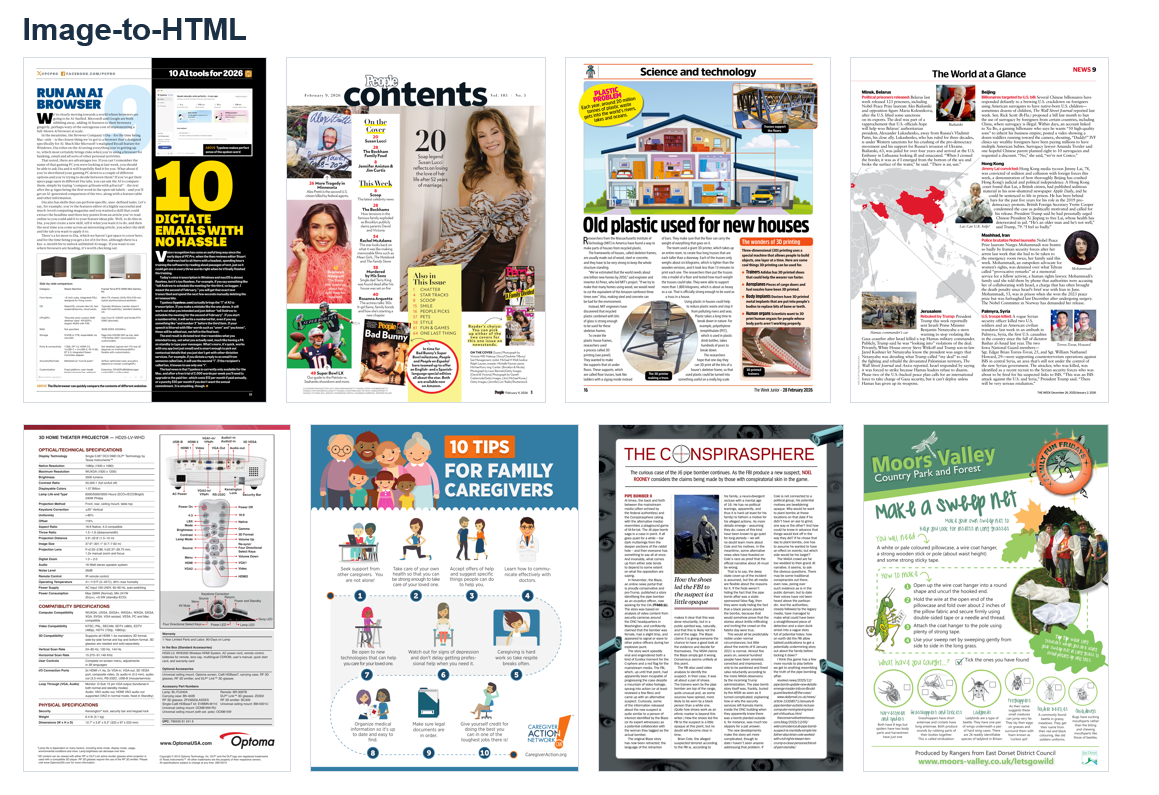}\par
    \vspace{4pt}
    \includegraphics[width=0.94\textwidth]{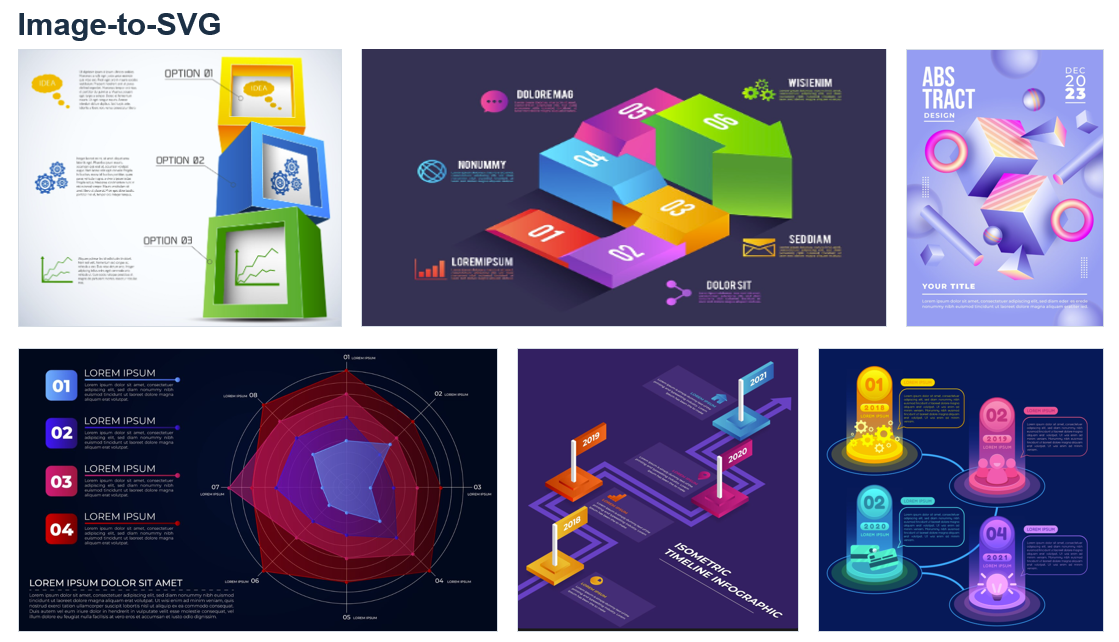}
    \caption{Representative benchmark test cases for Image-to-HTML (top) and Image-to-SVG (bottom).}
    \label{fig:appendix_case_illustration}
\end{figure*}

\begin{figure*}[!t]
    \centering
    \includegraphics[width=0.90\textwidth]{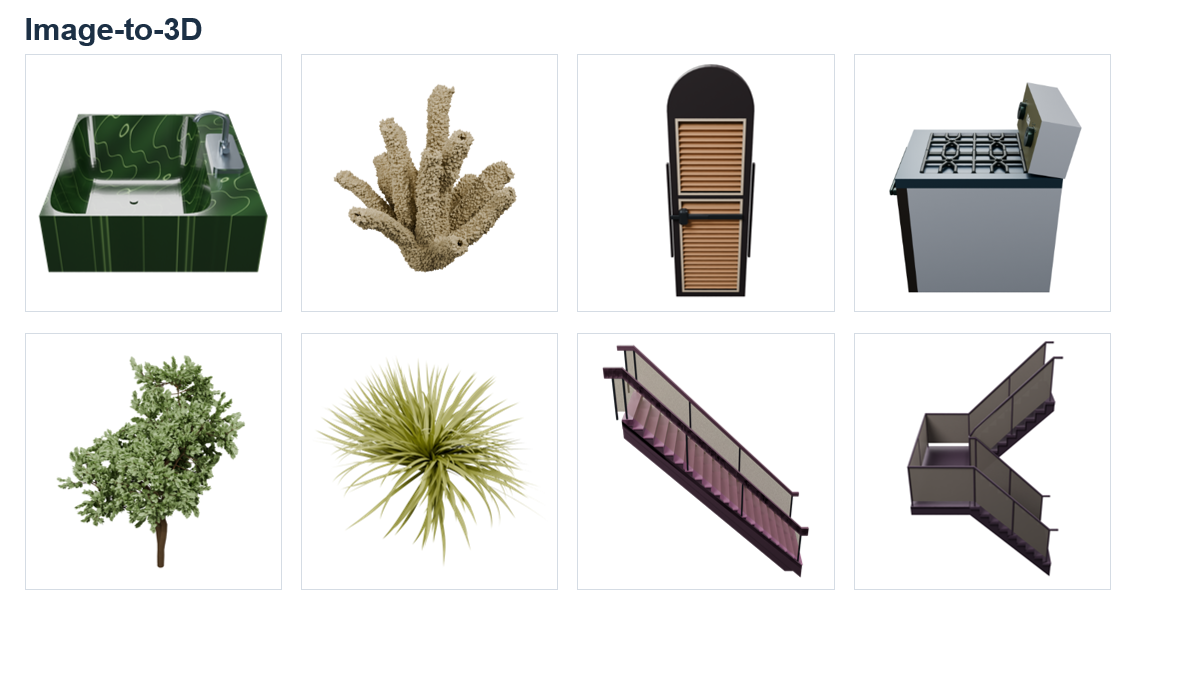}
    \caption{Representative benchmark test cases in 3DCodeBench.}
    \label{fig:appendix_case_3d}
    \vspace{10pt}
    \includegraphics[width=0.92\textwidth]{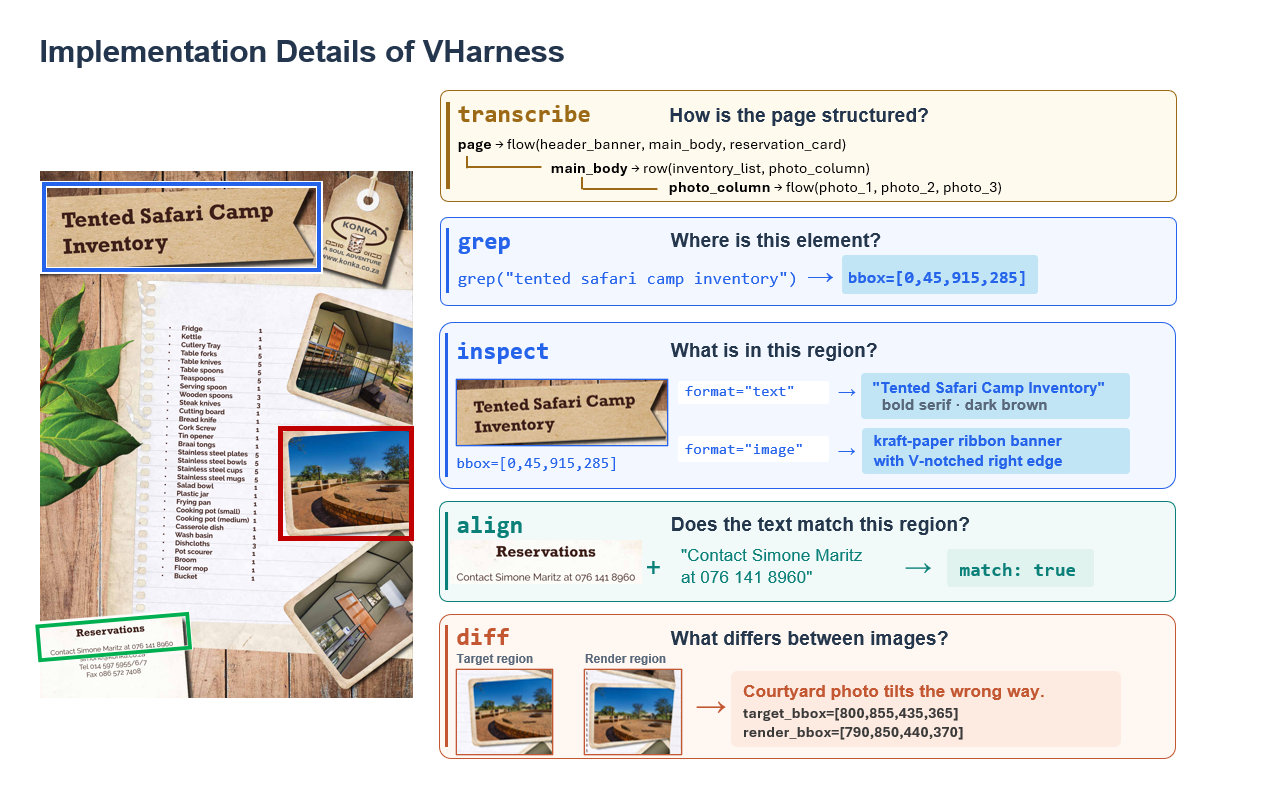}
    \caption{Input-output examples of the five atomic visual actions (\texttt{transcribe}, \texttt{grep}, \texttt{inspect}, \texttt{align}, and \texttt{diff}) implemented in VHarness.}
    \label{fig:appendix_impl_details}
\end{figure*}

On existing visual coding benchmarks such as Design2Code~\citep{si2025design2code}, which are dominated by simple grid templates or basic icons, modern MLLMs already achieve very high scores, leaving limited room to differentiate perceptual capabilities. To rigorously evaluate VHarness, we curate 100 high-complexity test cases each for Image-to-HTML and Image-to-SVG from diverse publicly available sources, paired with 100 official test cases from 3DCodeBench~\citep{gao2026codebench} for Image-to-3D (Figures~\ref{fig:appendix_case_illustration} and~\ref{fig:appendix_case_3d}). All experiments run in isolated Linux workspaces with Python 3.12, pre-installed scientific/vision libraries (\texttt{numpy}, \texttt{scipy}, \texttt{PIL}, \texttt{cv2}, \texttt{imageio}, \texttt{matplotlib}, \texttt{scikit-image}, \texttt{trimesh}), and on-demand PyPI package installation:

\begin{itemize}
    \item \textbf{Image-to-HTML.} To comprehensively evaluate layout reconstruction across diverse visual media, we collect 100 cases from open-source datasets and public repositories spanning multi-section webpages, editorial magazines, promotional posters, and presentation slides (Figure~\ref{fig:appendix_case_illustration}, top). Cases are selected by structural complexity, prioritizing non-trivial reading flows, multi-column asymmetric grids, dense typographic hierarchies, and irregular text wrapping. Non-textual visual assets (illustrations, photos, icons) are verified item-by-item through human inspection to ensure accurate cropping. Each workspace provides \texttt{target.png}, the verified assets under \texttt{./assets/}, and \texttt{provided\_assets.json} listing only filenames and native pixel dimensions with all spatial coordinates stripped (Figure~\ref{fig:appendix_visual_assets}). The agent authors a self-contained \texttt{code.html} with a single root container (\texttt{<div class="page">}), which Playwright/Chromium screenshots into \texttt{render.png}.
    \item \textbf{Image-to-SVG.} We collect 100 graphic designs from publicly available image and design platforms (e.g., Magnific and open-source vector/illustration galleries; Figure~\ref{fig:appendix_case_illustration}, bottom), filtering candidates by geometric and compositional complexity to retain multi-layered primitives, intricate B\'ezier contours, fine-grained local details, and rich color gradients. Each workspace provides only \texttt{target.png} without external assets; \texttt{<image>} tags, base64 data URIs, and traced bitmaps are strictly prohibited. The agent authors a pure-vector \texttt{code.svg} matching the target canvas, which headless Chromium rasterizes into \texttt{render.png}.
    \item \textbf{Image-to-3D.} We randomly sample 100 official test cases from 3DCodeBench~\citep{gao2026codebench} (Figure~\ref{fig:appendix_case_3d}) under the identical evaluation protocol to ensure strict benchmark consistency. Each workspace contains four canonical reference views (\texttt{target\_005/015/025/035.png} at azimuths $45^\circ, 135^\circ, 225^\circ, 315^\circ$ and $30^\circ$ elevation) and a read-only \texttt{render.py}. The agent authors a Blender 5.0 Python script (\texttt{code.py}) using \texttt{bpy}, \texttt{bmesh}, and \texttt{mathutils} with procedural \texttt{Principled BSDF} materials; running \texttt{python render.py} in headless Blender 5.0.1 renders the four matching views and exports the \texttt{.glb} mesh.
\end{itemize}

\subsection{3DCodeBench Ground-Truth Alignment}
\label{app:3d_alignment}

In the official release of 3DCodeBench~\citep{gao2026codebench}, the provided input images (\texttt{3DCodeBench/images}) do not match the corresponding \texttt{.glb} meshes and \texttt{.py} scripts in several cases, as documented in Issue~\#2 of the official repository. To ensure that the input reference views strictly correspond to the ground-truth 3D object, we use the open-source material Python scripts (\texttt{3DCodeData/<Name>\_000}) from the official Hugging Face dataset to generate the ground-truth 3D \texttt{.glb} artifacts in headless Blender 5.0.1, and capture the four canonical ground-truth views ($45^\circ$, $135^\circ$, $225^\circ$, and $315^\circ$) directly from those rendered 3D artifacts under the shared camera and lighting rig.

\subsection{VHarness and Implementation skills}
\label{app:mcp_sandbox}

\paragraph{FastMCP Server.}
VHarness is implemented as a persistent Model Context Protocol (MCP) server using \texttt{FastMCP}, exposing the five atomic visual actions (\texttt{transcribe}, \texttt{grep}, \texttt{inspect}, \texttt{align}, and \texttt{diff}) as callable tools under the \texttt{mcp\_\_atomic\_\_*} namespace.

\subsection{System Prompts and Agent Skills}
\label{app:prompts_skills}

We present the exact prompts and skill instructions used in our experiments: Figures~\ref{fig:prompts_atomic_part1} and~\ref{fig:prompts_atomic_part2} show the backend prompts for the five atomic visual actions in VHarness, while Figures~\ref{fig:skill_baseline_2d} provide the agent skill instructions across the three vision-to-code tasks and the \texttt{Freestyle} ablation.


\begin{figure*}[!t]
\centering

\vspace{4pt}
\begin{tcolorbox}[vhpromptbox, title={Atomic Action 1: \texttt{grep}}]
This is a \{w\}x\{h\}px image. Find EVERY element that matches this query: "\{query\}".\\
For each match return its category and a tight bounding box.\\
Use your standard bounding-box convention: [y\_min, x\_min, y\_max, x\_max] with each value NORMALIZED to the 0-1000 range (NOT pixels). Reply STRICT JSON array only:\\
{[}\{"category": "text|photo|logo|icon|chart|illustration|shape|other", "bbox": [y\_min, x\_min, y\_max, x\_max], "text": "<exact text if category=text, else empty>", "confidence": <0..1>\}{]}\\
Return [] if nothing matches.
\end{tcolorbox}

\vspace{4pt}
\begin{tcolorbox}[vhpromptbox, title={Atomic Action 2: \texttt{inspect}}]
\textbf{[format = "text"]}\\
Transcribe EXACTLY the text visible in this image, preserving line breaks and reading order. Then, on a NEW final line, report this text's TYPOGRAPHY, choosing from these fixed vocabularies:\\
\hspace*{1em}family: serif | slab-serif | sans-geometric | sans-humanist | sans-grotesque | monospace | display | script\\
\hspace*{1em}weight: thin | light | regular | medium | semibold | bold | black\\
\hspace*{1em}width:  condensed | normal | expanded \quad slant: italic | normal\\
and the text's INK colour as \#RRGGBB (the colour of the letters themselves, NOT the background). Format that final line EXACTLY as:\\
{[}type{]} family: <f> $\cdot$ weight: <w> $\cdot$ width: <wd> $\cdot$ slant: <s> $\cdot$ text-colour: \#RRGGBB\\
Output ONLY the transcribed text, then that single [type] line. No other commentary.\\[3pt]
\textbf{[format = "image"]}\\
Describe this visual element in detail: what it depicts, its colours, shapes, internal structure, and any embedded text. Be concrete.
\end{tcolorbox}
\caption{Exact backend prompts for the single-image perception actions (\texttt{grep}, and \texttt{inspect}).}
\label{fig:prompts_atomic_part1}
\end{figure*}

\begin{figure*}[!t]
\centering
\begin{tcolorbox}[vhpromptbox, title={Atomic Action 3: \texttt{align}}]
Does the following text match the text/content shown in this image region? Judge by meaning + wording, ignore font.\\
Reply STRICT JSON only: \{"match": true|false\}.\\
Text: \{text\}
\end{tcolorbox}

\begin{tcolorbox}[vhpromptbox, title={Atomic Action 4: \texttt{transcribe}}]
\textbf{[HTML / SVG]}\\
Your ONLY job is to recover the page SKELETON that the builder must write BEFORE inserting any text, image, number, or other content. Decide only: what the page's major regions are, in what order they read, which regions sit side by side, and how content flows inside each region. This is a real document structure, not an element-placement plan. You do NOT write code, and you do NOT output any coordinates, sizes, colours, or fonts.\\
Output the SKELETON as a STRUCTURE TREE in STRICT JSON, and nothing else:\\
\hspace*{1em}\{"layout": "flow|row|grid", "children": [ ... ]\}\\
where each entry in "children" is ONE of:\\
\hspace*{1em}- a LEAF referencing a located element by its id: \{"block": "block\_N"\} or \{"graphic": "asset\_NN"\}\\
\hspace*{1em}- a nested REGION CONTAINER: \{"layout": "flow|row|grid", "children": [ ... ]\} (a "grid" container ALSO carries "rows": <int> and "cols": <int>)\\[3pt]
\textbf{[3D]}\\
You are given \{n\} rendered views of ONE object on a turntable: azimuth \{az\} degrees, in that order, same camera height and distance. They are the SAME object seen from different sides --- a single view loses depth, so CROSS-CHECK the views to decide the real 3D shape (a circle in one view can be a sphere, a disc, or a cylinder seen end-on; the other views disambiguate it).
Your ONLY job is to recover the 3D object's structural decomposition before any code is written. Identify the major geometric parts, their primitive shapes, relative spatial arrangement, proportions, and observed surface colours/materials across views. You do NOT write code.
Output the 3D structure as a hierarchical JSON specification of parts, spatial relations, and material attributes.
\end{tcolorbox}

\vspace{4pt}
\begin{tcolorbox}[vhpromptbox, title={Atomic Action 5: \texttt{diff}}]
\textbf{[Prominent-Ranked Bounding-Box and Crop Comparison]}\\
Image 1 = the REFERENCE. Image 2 = the CANDIDATE reproduction of it. Report the ways the candidate DIFFERS from the reference, RANKED BY VISUAL PROMINENCE --- the order a viewer notices them: overall STRUCTURE / layout / architecture and the most prominent elements FIRST; finer, tail-end details LAST. Report macro before micro; do NOT let trivial micro details precede or crowd out the macro ones (you MAY still list micro differences, at the end). Cover: layout / column organisation, presence or absence of prominent elements, dominant / background colours, obvious size or prominence mismatches, text or label DOUBLING / GHOSTING, and overall faithfulness.\{focus\}\\
Reply STRICT JSON array only, MOST-PROMINENT FIRST: {[}\{"where":"<element / area>","what":"<the difference>","prominence":"high|medium|low"\}{]}. Return [] if the two match.\\[4pt]
\textbf{[Pairwise Bounding-Box and Crop Comparison]}\\
Image 1 = the TARGET. Image 2 = a REPRODUCTION of it (a render at the SAME canvas size). \{text\} shapes \& geometry, colours, layout \& proportions, object presence / completeness, and depth / occlusion. \{mode\} For EACH mismatch give a PAIR of boxes, because a misrendered element does NOT sit at the same coordinates in the two images:\\
\hspace*{1em}- "tag": 2-5 words naming the element (e.g. "blue cube", "pie chart", "tilted Step 03 label").\\
\hspace*{1em}- "target\_bbox": where that element is in the TARGET (image 1), [y\_min,x\_min,y\_max,x\_max] normalized 0-1000.\\
\hspace*{1em}- "render\_bbox": where the reproduction's attempt at it is in the RENDER (image 2), same format; use null if it is MISSING from the render.\\
Reply STRICT JSON array only: {[}\{"tag": "...", "target\_bbox": [y, x, y, x], "render\_bbox": [y, x, y, x]\}{]}. Return [] if the content already matches the target well.
\end{tcolorbox}
\caption{Exact backend prompts for the artifact verification actions (\texttt{align}, \texttt{transcribe} and \texttt{diff}).}
\label{fig:prompts_atomic_part2}
\end{figure*}



\begin{figure*}[!t]
\centering
\begin{tcolorbox}[vhpromptbox, title={Baseline Skill 1: Image-to-HTML (\texttt{w/o VHarness})}]
Reproduce the page shown in ./target.png as ONE self-contained HTML file ./code.html. Every graphic on the page is already provided in ./assets/ -- see provided\_assets.json for the file list and pixel sizes -- place them with <img> and size them with CSS.\\[3pt]
Do this hands-on: WRITE ./code.html directly with the Write tool and iterate -- do NOT present a plan or ask for approval; there is no one to approve it.\\[3pt]
You CAN SEE -- use Read on ./target.png to look at the reference.\\[3pt]
Work in TWO PASSES: FIRST Write the skeleton -- the <div class="page"> with empty containers for the page's regions, in reading order, nothing else. THEN fill each region with its content. Within a region the content flows: each text block and image follows the one before it -- do NOT put absolute positioning on a text block or an image itself. What you measure tells you which region something belongs to and how big it is, not where to put it on the page.\\[3pt]
In BOTH passes, all page content lives inside a single top-level <div class="page"> -- the renderer screenshots THAT element, so anything outside it is invisible -- at the reference's width; its height comes from its content. SELF-CONTAINED: inline <style>, no external CSS/JS, no web fonts. Build the page as a document, not as a canvas: a sequence of regions in reading order, each sized by what it holds. Reproduce layout, blocks, colours and TEXT (real text nodes: right content, size, weight, placement). Every readable text must appear exactly once and be opaque -- never hide text to fake a match. Take images only from ./assets/ -- do not crop or screenshot the reference, and do not download images; anything not in ./assets/ you may draw with CSS or generate with code.\\[3pt]
Then check yourself and ITERATE until it matches:\\
\hspace*{1em}1. run `python3 render.py` (Bash) -- it screenshots your .page to ./render.png at the reference's size. Use `python3` for any script you write, and keep every path relative to this workspace.\\
\hspace*{1em}2. Read ./render.png and compare it to ./target.png.\\
\hspace*{1em}3. fix ./code.html with Edit to close the gaps.\\
\hspace*{1em}4. repeat render -> compare -> refine until your render is as close to the reference as you can make it.\\
Then stop. The final ./code.html is your answer.
\end{tcolorbox}

\vspace{6pt}
\begin{tcolorbox}[vhpromptbox, title={Baseline Skill 2: Image-to-SVG (\texttt{w/o VHarness})}]
Reproduce the image shown in ./target.png as ONE self-contained, PURE-VECTOR SVG file ./code.svg.\\[3pt]
Do this hands-on: WRITE ./code.svg directly with the Write tool and iterate -- do NOT present a plan or ask for approval; there is no one to approve it.\\[3pt]
You CAN SEE -- use Read on ./target.png to look at the reference.\\[3pt]
Author ./code.svg with Write: a standalone `<svg>` document with an explicit width/height (or viewBox) matching the reference. PURE VECTOR ONLY -- no <image>, no embedded raster / base64, no external files or web fonts. Reproduce layout, shapes, strokes, fills, gradients and TEXT (as real <text> elements, not outlines).\\[3pt]
Then check yourself and ITERATE until it matches:\\
\hspace*{1em}1. run `python render.py` (Bash) -- it rasterizes your ./code.svg to ./render.png at the reference's size.\\
\hspace*{1em}2. Read ./render.png and compare it to ./target.png.\\
\hspace*{1em}3. fix ./code.svg with Edit to close the gaps (positions, sizes, colours, text content, font size/weight, spacing).\\
\hspace*{1em}4. repeat render -> compare -> refine until your render is as close to the reference as you can make it.\\
Then stop. The final ./code.svg is your answer.
\end{tcolorbox}
\caption{Exact skill instructions for the \texttt{w/o VHarness} baseline on Image-to-HTML and Image-to-SVG.}
\label{fig:skill_baseline_2d}
\end{figure*}

\clearpage


\begin{figure*}[!t]
\centering
\begin{tcolorbox}[vhpromptbox, title={Baseline Skill 3: Image-to-3D (\texttt{w/o VHarness})}]
Reproduce the 3D object shown in the 4 reference views as a single Blender 5.0 Python script ./code.py.\\[3pt]
Do this hands-on: WRITE ./code.py directly with the Write tool and iterate --- do NOT present a plan or ask for approval; there is no one to approve it.\\[3pt]
You CAN SEE --- use Read on ./target\_005.png, ./target\_015.png, ./target\_025.png and ./target\_035.png. They are the SAME object at four canonical turntable angles (azimuth 45 / 135 / 225 / 315 degrees, camera slightly above the object), rendered IN FULL COLOUR with their real materials.\\[3pt]
Author ./code.py with Write: a Blender 5.0 `bpy` script that builds the object(s) procedurally (primitives, meshes, modifiers, transforms). Leave the finished geometry in the scene --- do NOT add camera / light / render / export code (the harness supplies its own rig).\\[3pt]
Reproduce the MATERIALS / COLOUR too, not just the shape. Give each part a procedural Principled BSDF material with the colour you see in the reference, e.g.:\\
\hspace*{1.5em}m = bpy.data.materials.new('leaf'); m.use\_nodes = True\\
\hspace*{1.5em}b = m.node\_tree.nodes['Principled BSDF']\\
\hspace*{1.5em}b.inputs['Base Color'].default\_value = (0.30, 0.55, 0.15, 1) \quad \# r,g,b in 0..1\\
\hspace*{1.5em}b.inputs['Roughness'].default\_value = 0.6\\
\hspace*{1.5em}obj.data.materials.append(m)\\
Assign a DIFFERENT material per part (e.g. leaves vs trunk, seat vs legs) so multi-colour objects look right. No image textures / external files --- colour comes from materials only.\\[3pt]
Then check yourself and ITERATE until it matches:\\
\hspace*{1em}1. run `python render.py` (Bash) --- it renders your ./code.py to ./render\_005.png, ./render\_015.png, ./render\_025.png and ./render\_035.png (the SAME four angles, in colour).\\
\hspace*{1em}2. Read those 4 renders and compare each to the matching ./target\_0XX.png.\\
\hspace*{1em}3. fix ./code.py with Edit to close the gaps (proportions, parts, counts, orientation, and the base COLOUR of each part).\\
\hspace*{1em}4. repeat render -> compare -> refine until your render is as close to the reference as you can make it from all four angles, in both shape and colour.\\
Then stop. The final ./code.py is your answer.
\end{tcolorbox}


\caption{Exact skill instructions for the \texttt{w/o VHarness}
baseline on Image-to-3D.}
\label{fig:skill_baseline_3d}
\end{figure*}

\clearpage


\begin{figure*}[!t]
\centering
\begin{tcolorbox}[
    vhpromptbox,
    title={Ablation Skill: Image-to-HTML (\texttt{Freestyle})}
]
Reproduce the page shown in ./target.png as ONE self-contained HTML file ./code.html. Every graphic on the page is already provided in ./assets/ -- see provided\_assets.json for the file list and pixel sizes -- place them with \texttt{<img>} and size them with CSS.\\[3pt]
Do this hands-on: WRITE ./code.html directly with the Write tool and iterate -- do NOT present a plan or ask for approval; there is no one to approve it.\\[3pt]
You CAN SEE -- use Read on ./target.png to look at the reference.\\[3pt]
Work in TWO PASSES: FIRST Write the skeleton -- the
\texttt{\detokenize{<div class="page">}}
with empty containers for the page's regions, in reading order, nothing else. THEN fill each region with its content. Within a region the content flows: each text block and image follows the one before it -- do NOT put absolute positioning on a text block or an image itself. What you measure tells you which region something belongs to and how big it is, not where to put it on the page.\\[3pt]
In BOTH passes, all page content lives inside a single top-level
\texttt{\detokenize{<div class="page">}}
-- the renderer screenshots THAT element, so anything outside it is invisible -- at the reference's width; its height comes from its content. SELF-CONTAINED: inline \texttt{<style>}, no external CSS/JS, no web fonts. Build the page as a document, not as a canvas: a sequence of regions in reading order, each sized by what it holds. Reproduce layout, blocks, colours and TEXT (real text nodes: right content, size, weight, placement). Every readable text must appear exactly once and be opaque -- never hide text to fake a match. Take images only from ./assets/ -- do not crop or screenshot the reference, and do not download images; anything not in ./assets/ you may draw with CSS or generate with code.\\[3pt]
Then check yourself and ITERATE until it matches:\\
\hspace*{1em}1. run \texttt{python3 render.py} (Bash) -- it screenshots your .page to ./render.png at the reference's size. Use \texttt{python3} for any script you write, and keep every path relative to this workspace.\\
\hspace*{1em}2. Read ./render.png and compare it to ./target.png.\\
\hspace*{1em}3. fix ./code.html with Edit to close the gaps.\\
\hspace*{1em}4. repeat render -> compare -> refine until your render is as close to the reference as you can make it.\\
Then stop. The final ./code.html is your answer.\\[3pt]
\textbf{You also have access to a unified, general-purpose multimodal assistant (Gemini-3.8-flash) via the MCP tool \texttt{mcp\_\_raw\_mllm\_\_call}, which you may use as an auxiliary visual tool to help complete this task (e.g., observing visual details, estimating layout/geometry, and so on)\\[3pt]}
Pass \texttt{request} as an OpenAI-style Chat object:
\texttt{\detokenize{{"messages": [{"role": "user", "content": [{"type": "text", "text": "<your question>"}, {"type": "image_url", "image_url": {"url": "./target.png"}}]}], "max_tokens": 4096}}}.\\[3pt]
You may attach one or more workspace images in \texttt{content} using relative file paths (such as \texttt{./target.png}, \texttt{./render.png}, or files under \texttt{./assets/}; do NOT use \texttt{file://} URIs or external URLs).\\[3pt]
The tool executes your request as-is and returns the raw Chat completion envelope; you decide when and how to query it during your workflow.
\end{tcolorbox}
\caption{Complete skill instructions for the \texttt{Freestyle}
ablation on Image-to-HTML.}
\label{fig:skill_freestyle_html}
\end{figure*}

\clearpage


\begin{figure*}[!t]
\centering
\begin{tcolorbox}[
    vhpromptbox,
    title={Ablation Skill: Image-to-3D (\texttt{Freestyle})}
]
Reproduce the 3D object shown in the 4 reference views as a single Blender 5.0 Python script ./code.py.\\[3pt]
Do this hands-on: WRITE ./code.py directly with the Write tool and iterate --- do NOT present a plan or ask for approval; there is no one to approve it.\\[3pt]
You CAN SEE --- use Read on ./target\_005.png, ./target\_015.png, ./target\_025.png and ./target\_035.png. They are the SAME object at four canonical turntable angles (azimuth 45 / 135 / 225 / 315 degrees, camera slightly above the object), rendered IN FULL COLOUR with their real materials.\\[3pt]
Author ./code.py with Write: a Blender 5.0 \texttt{bpy} script that builds the object(s) procedurally (primitives, meshes, modifiers, transforms). Leave the finished geometry in the scene --- do NOT add camera / light / render / export code (the harness supplies its own rig).\\[3pt]
Reproduce the MATERIALS / COLOUR too, not just the shape. Give each part a procedural Principled BSDF material with the colour you see in the reference, e.g.:\\
\hspace*{1.5em}m = bpy.data.materials.new('leaf'); m.use\_nodes = True\\
\hspace*{1.5em}b = m.node\_tree.nodes['Principled BSDF']\\
\hspace*{1.5em}b.inputs['Base Color'].default\_value = (0.30, 0.55, 0.15, 1) \quad \# r,g,b in 0..1\\
\hspace*{1.5em}b.inputs['Roughness'].default\_value = 0.6\\
\hspace*{1.5em}obj.data.materials.append(m)\\
Assign a DIFFERENT material per part (e.g. leaves vs trunk, seat vs legs) so multi-colour objects look right. No image textures / external files --- colour comes from materials only.\\[3pt]
Then check yourself and ITERATE until it matches:\\
\hspace*{1em}1. run \texttt{python render.py} (Bash) --- it renders your ./code.py to ./render\_005.png, ./render\_015.png, ./render\_025.png and ./render\_035.png (the SAME four angles, in colour).\\
\hspace*{1em}2. Read those 4 renders and compare each to the matching ./target\_0XX.png.\\
\hspace*{1em}3. fix ./code.py with Edit to close the gaps (proportions, parts, counts, orientation, and the base COLOUR of each part).\\
\hspace*{1em}4. repeat render -> compare -> refine until your render is as close to the reference as you can make it from all four angles, in both shape and colour.\\
Then stop. The final ./code.py is your answer.\\[3pt]
\textbf{You also have access to a unified, general-purpose multimodal assistant (Gemini-3.8-flash) via the MCP tool \texttt{mcp\_\_raw\_mllm\_\_call}, which you may use as an auxiliary visual tool to help complete this task (e.g., observing visual details, estimating layout/geometry, and so on)\\[3pt]}
Pass \texttt{request} as an OpenAI-style Chat object:
\texttt{\detokenize{{"messages": [{"role": "user", "content": [{"type": "text", "text": "<your question>"}, {"type": "image_url", "image_url": {"url": "./target.png"}}]}], "max_tokens": 4096}}}.\\[3pt]
You may attach one or more workspace images in \texttt{content} using relative file paths (such as \texttt{./target.png}, \texttt{./render.png}, or files under \texttt{./assets/}; do NOT use \texttt{file://} URIs or external URLs).\\[3pt]
The tool executes your request as-is and returns the raw Chat completion envelope; you decide when and how to query it during your workflow.
\end{tcolorbox}
\caption{Complete skill instructions for the \texttt{Freestyle}
ablation on Image-to-3D.}
\label{fig:skill_freestyle_3d}
\end{figure*}

\clearpage

\clearpage
\section{Detailed Experimental Setup}
\label{app:experimental_setup}

\subsection{Model and Hyperparameter Configurations}
\label{app:hyperparameters}

To ensure a fair comparison across all tasks and model families, all coding agents are executed within the same Claude Code CLI harness with a fixed context window of 200K tokens (with automatic context compaction enabled for long multi-turn trajectories). Table~\ref{tab:app_model_configs} summarizes the model configurations and inference hyperparameters used in our main experiments and same-model ablation study. 

Due to inference budget constraints, we evaluate GPT-6-astra with \texttt{low} reasoning effort, rather than the \texttt{high} setting used in the main experiments. Despite using \texttt{low} reasoning effort, the GPT-6-astra baseline achieves better scores than every configuration in Table~\ref{tab:main_results}(b) on all metrics reported in Table~\ref{tab:astra_same_model}, indicating strong visual-coding performance even under this reduced-reasoning setting. The primary purpose of the experiment is to compare GPT-6-astra with and without VHarness to verify whether
the Vision-as-a-Tool (VaaT) paradigm stays effective as a model’s intrinsic vision becomes stronger.

\begin{table*}[!ht]
    \centering
    \caption{Summary of model configurations and inference hyperparameters across the main experiments and ablation study. All coding agents operate under a fixed 200K context window.}
    \label{tab:app_model_configs}
    \small
    \setlength{\tabcolsep}{6pt}
    \renewcommand{\arraystretch}{1.15}
    \begin{tabular*}{\textwidth}{@{\extracolsep{\fill}}lccc@{}}
        \toprule
        Model & Context Window & Reasoning Effort \\
        \midrule
        GPT-5.5~\citep{openai2026gpt55} & 200K & \texttt{high} \\
        GPT-5.6-sol~\citep{openai2026gpt56} & 200K & \texttt{high} \\
        Claude Opus 4.8~\citep{anthropic2026opus48} & 200K & \texttt{high} \\
        GLM-5.2~\citep{zai2026glm52} & 200K & \texttt{high} \\
        GLM-5.3-Flash~\citep{zai2026glm53flash} & 200K & \texttt{high}  \\
        GPT-6-astra (Ablation) & 200K & \texttt{low} \\
        \midrule
        Gemini-3.8-Flash ($\mathcal{U}_{\phi}$) & 200K & \texttt{high} \\
        \bottomrule
    \end{tabular*}
\end{table*}

\subsection{Evaluation Metric Details}
\label{app:evaluation_metrics}

\paragraph{2D Perceptual Fidelity Metrics.}
We evaluate semantic and perceptual alignment using three complementary metrics:
\begin{itemize}
    \item \textbf{SigLIP2 ($\uparrow$).} We extract dense patch embeddings using \texttt{google/siglip2-so400m-patch16-\allowbreak naflex}. To evaluate high-resolution webpages and graphics without downsampling distortion, both the target and rendered images are partitioned into overlapping $448 \times 448$ tiles with a stride of $224$ pixels. Overlapping patch observations are de-duplicated at the global pixel grid, weighted by valid non-padded pixel area, and compared via cosine similarity.
    \item \textbf{DINOv2 ($\uparrow$).} We extract tile-level \texttt{[CLS]} feature vectors using \texttt{facebook/dinov2-large}~\citep{oquab2023dinov2} over overlapping $448 \times 448$ tiles with a stride of $224$ pixels at native resolution, and compute the mean cosine similarity across all corresponding tile pairs.
    \item \textbf{LPIPS ($\downarrow$).} We compute the Learned Perceptual Image Patch Similarity distance~\citep{zhang2018lpips} using the AlexNet backbone (\texttt{net\_type="alex"}) with input RGB tensors normalized to $[0, 1]$ (\texttt{normalize=True}). Lower values indicate higher perceptual similarity.
\end{itemize}

\paragraph{2D Structural Fidelity Metrics.}
We measure spatial layout, pixel accuracy, and contour alignment using four structural metrics:
\begin{itemize}
    \item \textbf{Local SSIM (\texttt{SSIM l.}, $\uparrow$).} We compute the standard windowed Structural Similarity Index~\citep{wang2004ssim} across RGB channels at native resolution with \texttt{data\_range=255}, measuring local luminance, contrast, and structural consistency.
    \item \textbf{Global SSIM (\texttt{SSIM g.}, $\uparrow$).} To capture whole-canvas luminance and contrast consistency across large layout blocks, we compute the SSIM formula over the full-image mean, variance, and covariance statistics per RGB channel and average the resulting scores across the three channels.
    \item \textbf{Pixel MSE (\texttt{Px MSE}, $\downarrow$).} We normalize RGB pixel intensities to $[0, 1]$ by dividing by $255$ and compute the mean squared error across all pixels and color channels.
    \item \textbf{Canny Edge Precision, Recall, and F1 (\texttt{Edge P / R / F1}, $\uparrow$).} For Image-to-HTML and Image-to-SVG, we convert both the target and rendered images to grayscale and extract structural contours using a Canny edge detector. To account for minor sub-pixel or anti-aliasing shifts, predicted and reference edge maps are matched within a small spatial neighborhood tolerance to compute Edge Precision, Edge Recall, and their harmonic mean (Edge F1).
\end{itemize}

\paragraph{3D Geometric Accuracy Metrics.}
Following the official evaluation protocol of 3DCodeBench~\citep{gao2026codebench}, we evaluate the exported \texttt{.glb} mesh against the ground-truth mesh by deterministically sampling $N = 8{,}192$ surface points (and corresponding surface RGB colors) from each mesh and normalizing each point cloud into the unit sphere by subtracting its centroid and dividing by its maximum Euclidean radius:
\begin{itemize}
    \item \textbf{Uni3D ($\uparrow$).} We encode the normalized $8{,}192$-point clouds (concatenating 3D coordinates and RGB values into 6D point features) using the pretrained \texttt{Uni3D-giant} point-cloud encoder (\texttt{eva\_giant\_patch14\_560} backbone, $1{,}024$-dimensional embedding)~\citep{zhou2023uni3d} and compute the cosine similarity between the predicted and ground-truth 3D embeddings.
    \item \textbf{Chamfer Distance ($\downarrow$).} We compute the bidirectional squared Chamfer distance~\citep{fan2017pointset} between the unit-sphere normalized point clouds $P$ and $Q$:
    \begin{equation}
    \begin{aligned}
        \text{CD}(P, Q) &= \frac{1}{|P|}\sum_{p \in P} \min_{q \in Q} \|p - q\|_2^2 \\
        &\quad + \frac{1}{|Q|}\sum_{q \in Q} \min_{p \in P} \|q - p\|_2^2,
    \end{aligned}
    \end{equation}
    evaluated across the four canonical yaw rotations $\{0^\circ, 90^\circ, 180^\circ, 270^\circ\}$ and taking the minimum distance, matching the official 3DCodeBench shape scorer.
\end{itemize}

\section{Computational Cost, Same-Subset Evaluation, and Trajectory Analysis}
\label{app:cost_and_cases}

\subsection{Token Consumption and Action Invocation Statistics}
\label{app:cost_analysis}

\begin{table*}[!t]
    \centering
    \caption{Average recorded output tokens per case by the coding agent $\mathcal{P}_{\theta}$ across Image-to-HTML, Image-to-SVG, and Image-to-3D.}
    \label{tab:token_cost}
    \small
    \setlength{\tabcolsep}{4.5pt}
    \renewcommand{\arraystretch}{1.15}
    \begin{tabular*}{\textwidth}{@{\extracolsep{\fill}}lcrrr@{}}
        \toprule
        Model & VHarness & HTML & SVG & 3D \\
        \midrule
        GPT-5.5         & \texttt{w/o} &  52,792 &  30,983 &  37,886 \\
                        & \texttt{w/}  & 173,374 & 195,148 & 237,750 \\
        \addlinespace[2pt]
        GPT-5.6-sol     & \texttt{w/o} &  23,825 &  16,301 &  17,918 \\
                        & \texttt{w/}  &  62,005 &  64,841 & 104,543 \\
        \addlinespace[2pt]
        Claude Opus 4.8 & \texttt{w/o} &  14,132 &  14,153 &  14,522 \\
                        & \texttt{w/}  &  62,859 &  82,688 &  92,321 \\
        \addlinespace[2pt]
        GLM-5.2         & \texttt{w/}  & 685,988 & 1,971,312 & 164,741 \\
        \addlinespace[2pt]
        GLM-5.3-Flash   & \texttt{w/o} & 148,404 & 198,996 & 205,156 \\
                        & \texttt{w/}  & 291,953 & 433,949 & 619,680 \\
        \bottomrule
    \end{tabular*}
\end{table*}

Table~\ref{tab:token_cost} reports the average output token consumption per case by the coding agent $\mathcal{P}_{\theta}$. Importantly, neither \texttt{w/o VHarness} nor \texttt{w/ VHarness} imposes a minimum iteration count or forced token budget; both terminate once the rendered artifact is judged visually complete. Under \texttt{w/o VHarness}, stopping relies solely on intrinsic vision, which frequently overlooks fine-grained layout shifts, missing vector contours, or 3D geometric misalignments and terminates prematurely. Under \texttt{w/ VHarness}, external visual utilities continuously surface remaining discrepancies via \texttt{diff} and \texttt{inspect}, driving sustained refinement.

\begin{figure*}[!t]
    \centering
    \begin{subfigure}{\textwidth}
        \centering
        \includegraphics[width=\linewidth]{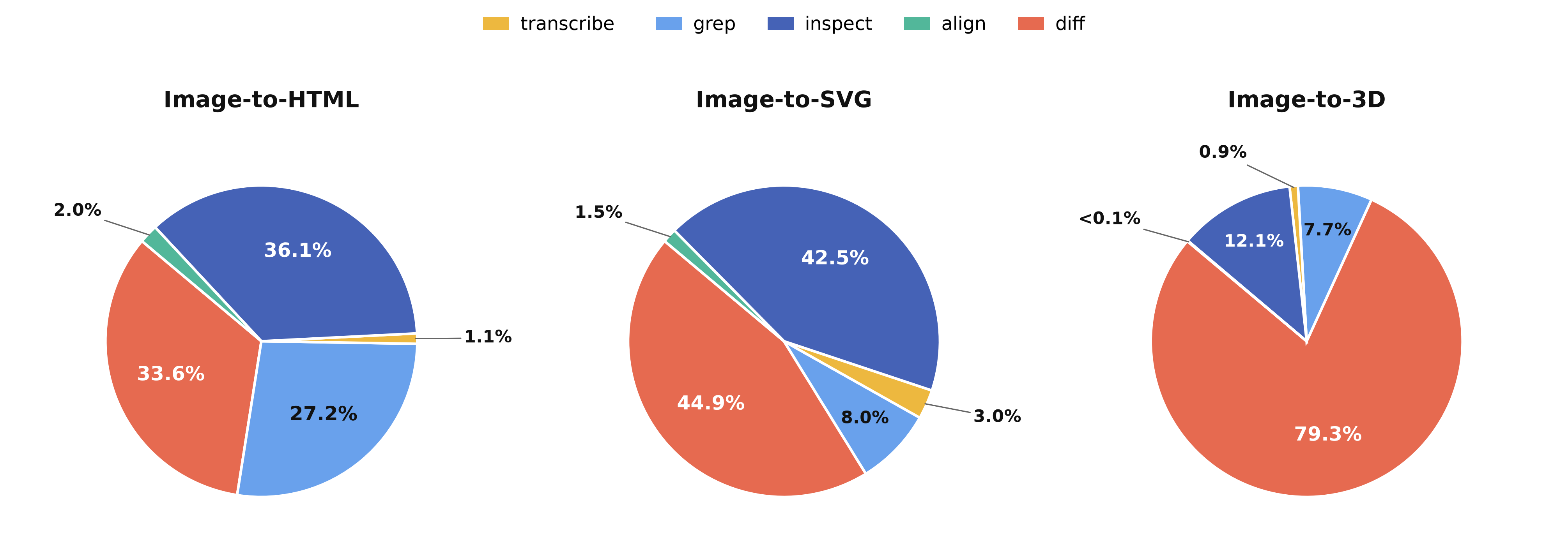}
        \caption{GLM-5.2: text-only coding agent.}
        \label{fig:action_glm52}
    \end{subfigure}
    \caption{\textbf{Atomic action invocation distributions across tasks.}
    Proportions are computed by normalizing
    the average per-case call counts across actions.}
    \label{fig:appendix_action_analysis}
\end{figure*}

Table~\ref{tab:action_counts} and Figure~\ref{fig:appendix_action_analysis} break down the atomic action invocations of the text-only agent GLM-5.2 across tasks, while Table~\ref{tab:tool_comparison} compares native harness tool calls against total VHarness actions across all evaluated models. 

\begin{table*}[!t]
    \centering
    \caption{Average number of atomic visual action calls per case invoked by the text-only coding agent GLM-5.2 under VHarness across Image-to-HTML, Image-to-SVG, and Image-to-3D.}
    \label{tab:action_counts}
    \small
    \setlength{\tabcolsep}{6pt}
    \renewcommand{\arraystretch}{1.15}
    \begin{tabular*}{\textwidth}{@{\extracolsep{\fill}}lrrrrrr@{}}
        \toprule
        Task & \texttt{transcribe} & \texttt{grep} & \texttt{inspect} & \texttt{align} & \texttt{diff} & Total \\
        \midrule
        Image-to-HTML &  7.8 & 199.5 & 264.8 & 14.3 & 246.5 & 733.0 \\
        Image-to-SVG  & 11.2 &  29.4 & 155.9 &  5.4 & 164.5 & 366.4 \\
        Image-to-3D   &  3.8 &  34.3 &  53.5 &  0.3 & 351.8 & 443.8 \\
        \bottomrule
    \end{tabular*}
\end{table*}

\begin{table*}[!t]
    \centering
    \caption{Average per-case invocations of native harness tools (\texttt{Read}, \texttt{Bash}, \texttt{Write}, \texttt{Edit}, and \texttt{Other}) and total VHarness atomic visual actions across Image-to-HTML, Image-to-SVG, and Image-to-3D. \texttt{Other} includes task-management and MCP resource tools.}
    \label{tab:tool_comparison}
    \small
    \setlength{\tabcolsep}{6pt}
    \renewcommand{\arraystretch}{1.15}
    \begin{tabular*}{\textwidth}{@{\extracolsep{\fill}}lrrrrrr@{}}
        \toprule
        \multicolumn{7}{@{}l}{\textbf{(a) Image-to-HTML}} \\
        Model & \texttt{Read} & \texttt{Bash} & \texttt{Write} & \texttt{Edit} & \texttt{Other} & \texttt{VHarness} \\
        \midrule
        GPT-5.5         &  72.5 &   58.3 &   3.3 &  74.7 & 13.6 & 187.4 \\
        GPT-5.6-sol     &  39.1 &   46.4 &   2.4 &  64.8 & 10.1 & 103.0 \\
        Claude Opus 4.8 &  48.7 &   37.6 &   3.2 &  43.3 &  0.0 &  87.4 \\
        GLM-5.2         &   0.0 & 1091.7 &  46.5 & 361.4 & 26.6 & 733.0 \\
        GLM-5.3-Flash   &  23.7 &  179.7 &  27.8 &  32.7 &  0.2 &  33.1 \\
        \midrule
        \multicolumn{7}{@{}l}{\textbf{(b) Image-to-SVG}} \\
        Model & \texttt{Read} & \texttt{Bash} & \texttt{Write} & \texttt{Edit} & \texttt{Other} & \texttt{VHarness} \\
        \midrule
        GPT-5.5         & 161.9 &   85.7 &   3.7 &  63.5 & 19.3 &  91.7 \\
        GPT-5.6-sol     &  64.7 &   47.7 &   2.3 &  54.1 & 13.3 &  56.4 \\
        Claude Opus 4.8 &  84.9 &   36.3 &   2.9 &  45.2 &  0.0 &  56.6 \\
        GLM-5.2         &   0.0 & 2490.7 & 227.7 & 221.3 & 54.2 & 366.4 \\
        GLM-5.3-Flash   & 117.6 &  250.1 &  33.1 &  40.2 &  0.7 &  11.1 \\
        \midrule
        \multicolumn{7}{@{}l}{\textbf{(c) Image-to-3D}} \\
        Model & \texttt{Read} & \texttt{Bash} & \texttt{Write} & \texttt{Edit} & \texttt{Other} & \texttt{VHarness} \\
        \midrule
        GPT-5.5         & 139.2 &  110.5 &   2.1 &  80.8 & 27.2 & 284.6 \\
        GPT-5.6-sol     & 104.0 &  106.4 &   1.5 & 114.2 & 24.0 & 242.6 \\
        Claude Opus 4.8 &  87.3 &   53.4 &   1.6 &  73.4 &  0.1 & 164.3 \\
        GLM-5.2         &   0.0 &  187.7 &   3.1 &  49.9 & 22.0 & 443.8 \\
        GLM-5.3-Flash   & 157.3 &  408.9 &  19.0 &  18.8 &  0.4 &  36.2 \\
        \bottomrule
    \end{tabular*}
\end{table*}

\subsection{Evaluation on the Shared 50-Case Subset}
\label{app:subset50_eval}

In Table~\ref{tab:main_results} of the main paper, the text-only agent GLM-5.2 is evaluated on a random 50-case subset per task due to inference budget constraints, whereas multimodal agents are evaluated on all 100 cases. To provide a strictly aligned comparison on the exact same test instances, Table~\ref{tab:original50_gpt_glm52} reports the results of GPT-5.5, GPT-5.6-sol, and GLM-5.2 on this shared 50-case subset, where all relative rankings and conclusions from the main text hold consistently.

\begin{table*}[!t]
    \centering
    \caption{Quantitative comparison on the same 50 cases evaluation testset.
    GPT-5.5 and GPT-5.6-sol are evaluated with (\texttt{w/}) and without
    (\texttt{w/o}) VHarness on all 50 cases per task. Arrows indicate the preferred direction for each metric, and bold marks the best displayed result in each column.}
    \label{tab:original50_gpt_glm52}

    \newcommand{\appbest}[1]{\textbf{#1}}
    \newcommand{\appmc}[1]{\makebox[0.08666\textwidth][c]{#1}}
    \newcommand{\appgc}[1]{\makebox[0.13\textwidth][c]{#1}}
    \newcommand{\appmodcol}[1]{\makebox[0.15\textwidth][l]{#1}}
    \newcommand{\appvhcol}[1]{\makebox[0.07\textwidth][c]{#1}}

    \footnotesize
    \setlength{\tabcolsep}{0pt}
    \renewcommand{\arraystretch}{1.16}

    \begin{tabular*}{\textwidth}{@{}llccccccccc@{}}
        \toprule
        \multicolumn{11}{@{}l}{\textbf{(a) Image-to-HTML}} \\
        \appmodcol{Model} & \appvhcol{VHarness}
        & \multicolumn{3}{c}{2D Perceptual}
        & \multicolumn{6}{c}{2D Structural} \\
        \cmidrule(lr){3-5}\cmidrule(l){6-11}
        & &
        \appmc{SigLIP2 $\uparrow$} &
        \appmc{DINOv2 $\uparrow$} &
        \appmc{LPIPS $\downarrow$} &
        \appmc{SSIM l. $\uparrow$} &
        \appmc{SSIM g. $\uparrow$} &
        \appmc{Px MSE $\downarrow$} &
        \appmc{Edge P $\uparrow$} &
        \appmc{Edge R $\uparrow$} &
        \appmc{Edge F1 $\uparrow$} \\
        \midrule

        \appmodcol{GPT-5.5} & \appvhcol{w/o}
        & \appmc{0.5973} & \appmc{0.8939} & \appmc{0.2129}
        & \appmc{0.6757} & \appmc{0.7122} & \appmc{0.0500}
        & \appmc{0.5817} & \appmc{0.5637} & \appmc{0.5717} \\
         & \appvhcol{w/}
        & \appmc{0.6400} & \appmc{0.9074} & \appmc{0.1807}
        & \appmc{0.6911} & \appmc{0.7284} & \appmc{0.0476}
        & \appmc{0.6169} & \appmc{0.5996} & \appmc{0.6073} \\
        \addlinespace[1.6pt]

        \appmodcol{GPT-5.6-sol} & \appvhcol{w/o}
        & \appmc{0.6459} & \appmc{0.9074} & \appmc{0.1694}
        & \appmc{0.6971} & \appmc{0.7453} & \appmc{0.0448}
        & \appmc{0.6526} & \appmc{0.6317} & \appmc{0.6410} \\
         & \appvhcol{w/}
        & \appmc{\appbest{0.6871}}
        & \appmc{\appbest{0.9128}}
        & \appmc{\appbest{0.1452}}
        & \appmc{\appbest{0.7265}}
        & \appmc{\appbest{0.7657}}
        & \appmc{\appbest{0.0412}}
        & \appmc{\appbest{0.7062}}
        & \appmc{\appbest{0.6876}}
        & \appmc{\appbest{0.6960}} \\
        \addlinespace[1.6pt]

        \appmodcol{GLM-5.2} & \appvhcol{w/}
        & \appmc{0.6044} & \appmc{0.8949} & \appmc{0.2028}
        & \appmc{0.6715} & \appmc{0.7242} & \appmc{0.0477}
        & \appmc{0.6107} & \appmc{0.5736} & \appmc{0.5907} \\

        \bottomrule
    \end{tabular*}

    \par\vspace{0.6em}

    \begin{tabular*}{\textwidth}{@{}llccccccccc@{}}
        \toprule
        \multicolumn{11}{@{}l}{\textbf{(b) Image-to-SVG}} \\
        \appmodcol{Model} & \appvhcol{VHarness}
        & \multicolumn{3}{c}{2D Perceptual}
        & \multicolumn{6}{c}{2D Structural} \\
        \cmidrule(lr){3-5}\cmidrule(l){6-11}
        & &
        \appmc{SigLIP2 $\uparrow$} &
        \appmc{DINOv2 $\uparrow$} &
        \appmc{LPIPS $\downarrow$} &
        \appmc{SSIM l. $\uparrow$} &
        \appmc{SSIM g. $\uparrow$} &
        \appmc{Px MSE $\downarrow$} &
        \appmc{Edge P $\uparrow$} &
        \appmc{Edge R $\uparrow$} &
        \appmc{Edge F1 $\uparrow$} \\
        \midrule

        \appmodcol{GPT-5.5} & \appvhcol{w/o}
        & \appmc{0.6304} & \appmc{0.8611} & \appmc{0.1546}
        & \appmc{0.7979} & \appmc{0.8137} & \appmc{0.0156}
        & \appmc{0.6550} & \appmc{0.6312} & \appmc{0.6417} \\
         & \appvhcol{w/}
        & \appmc{0.6553} & \appmc{0.8874} & \appmc{0.1248}
        & \appmc{0.8126} & \appmc{0.8445} & \appmc{0.0131}
        & \appmc{0.6815} & \appmc{0.6633} & \appmc{0.6712} \\
        \addlinespace[1.6pt]

        \appmodcol{GPT-5.6-sol} & \appvhcol{w/o}
        & \appmc{0.6682} & \appmc{0.8911} & \appmc{0.1137}
        & \appmc{0.8238} & \appmc{0.8571} & \appmc{0.0122}
        & \appmc{0.7169} & \appmc{0.7057} & \appmc{0.7102} \\
         & \appvhcol{w/}
        & \appmc{\appbest{0.6838}}
        & \appmc{\appbest{0.9064}}
        & \appmc{\appbest{0.1025}}
        & \appmc{0.8302}
        & \appmc{0.8631}
        & \appmc{0.0113}
        & \appmc{0.7194}
        & \appmc{\appbest{0.7076}}
        & \appmc{\appbest{0.7126}} \\
        \addlinespace[1.6pt]

        \appmodcol{GLM-5.2} & \appvhcol{w/}
        & \appmc{0.6507}
        & \appmc{0.8656}
        & \appmc{0.1219}
        & \appmc{\appbest{0.8471}}
        & \appmc{\appbest{0.8756}}
        & \appmc{\appbest{0.0094}}
        & \appmc{\appbest{0.7383}}
        & \appmc{0.6665}
        & \appmc{0.6950} \\

        \bottomrule
    \end{tabular*}

    \par\vspace{0.6em}

    \begin{tabular*}{\textwidth}{@{}llcccccccc@{}}
        \toprule
        \multicolumn{10}{@{}l}{\textbf{(c) Image-to-3D}} \\
        \appmodcol{Model} & \appvhcol{VHarness}
        & \multicolumn{3}{c}{2D Perceptual}
        & \multicolumn{3}{c}{2D Structural}
        & \multicolumn{2}{c}{3D Geometric} \\
        \cmidrule(lr){3-5}\cmidrule(lr){6-8}\cmidrule(l){9-10}
        & &
        \appmc{SigLIP2 $\uparrow$} &
        \appmc{DINOv2 $\uparrow$} &
        \appmc{LPIPS $\downarrow$} &
        \appmc{SSIM l. $\uparrow$} &
        \appmc{SSIM g. $\uparrow$} &
        \appmc{Px MSE $\downarrow$} &
        \appgc{Uni3D $\uparrow$} &
        \appgc{Chamfer $\downarrow$} \\
        \midrule

        \appmodcol{GPT-5.5} & \appvhcol{w/o}
        & \appmc{0.6133} & \appmc{0.6696} & \appmc{0.1470}
        & \appmc{0.8875} & \appmc{0.7048} & \appmc{0.0282}
        & \appgc{0.7666} & \appgc{0.0100} \\
         & \appvhcol{w/}
        & \appmc{0.6938} & \appmc{0.7659} & \appmc{0.0806}
        & \appmc{0.9204} & \appmc{0.8651} & \appmc{0.0080}
        & \appgc{0.7870} & \appgc{0.0049} \\
        \addlinespace[1.6pt]

        \appmodcol{GPT-5.6-sol} & \appvhcol{w/o}
        & \appmc{0.6435} & \appmc{0.7034} & \appmc{0.1178}
        & \appmc{0.9041} & \appmc{0.7397} & \appmc{0.0164}
        & \appgc{0.7681} & \appgc{0.0172} \\
         & \appvhcol{w/}
        & \appmc{\appbest{0.7263}}
        & \appmc{\appbest{0.8101}}
        & \appmc{\appbest{0.0625}}
        & \appmc{\appbest{0.9293}}
        & \appmc{\appbest{0.8963}}
        & \appmc{\appbest{0.0053}}
        & \appgc{\appbest{0.8383}}
        & \appgc{\appbest{0.0034}} \\
        \addlinespace[1.6pt]

        \appmodcol{GLM-5.2} & \appvhcol{w/}
        & \appmc{0.6400} & \appmc{0.6623} & \appmc{0.1250}
        & \appmc{0.9076} & \appmc{0.7470} & \appmc{0.0162}
        & \appgc{0.7548} & \appgc{0.0092} \\

        \bottomrule
    \end{tabular*}
\end{table*}

\subsection{Quantitative Refinement Dynamics on Image-to-3D}
\label{app:refinement_dynamics}

\begin{figure*}[!t]
    \centering
    \begin{subfigure}[t]{0.48\textwidth}
        \centering
        \includegraphics[width=\linewidth]{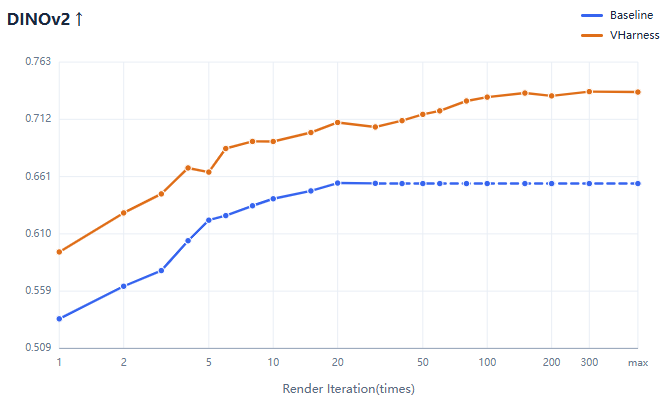}
        \caption{GPT-5.5 on DINOv2 ($\uparrow$).}
        \label{fig:gpt55_dino_3d}
    \end{subfigure}
    \hfill
    \begin{subfigure}[t]{0.48\textwidth}
        \centering
        \includegraphics[width=\linewidth]{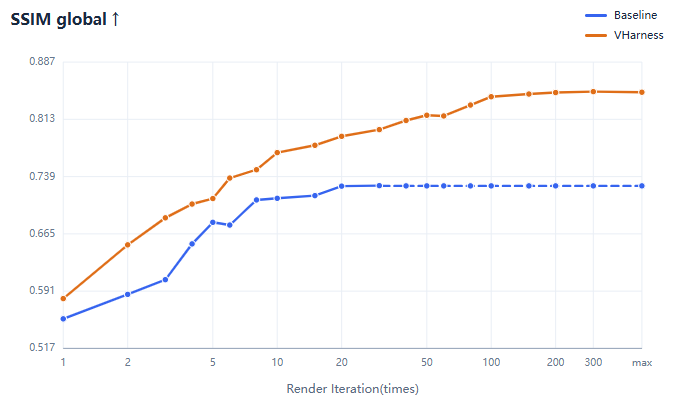}
        \caption{GPT-5.5 on global SSIM ($\uparrow$).}
        \label{fig:gpt55_ssim_3d}
    \end{subfigure}

    \vspace{6pt}

    \begin{subfigure}[t]{0.48\textwidth}
        \centering
        \includegraphics[width=\linewidth]{figures/gpt56-dino-3d.png}
        \caption{GPT-5.6-sol on DINOv2 ($\uparrow$).}
        \label{fig:gpt56_dino_3d}
    \end{subfigure}
    \hfill
    \begin{subfigure}[t]{0.48\textwidth}
        \centering
        \includegraphics[width=\linewidth]{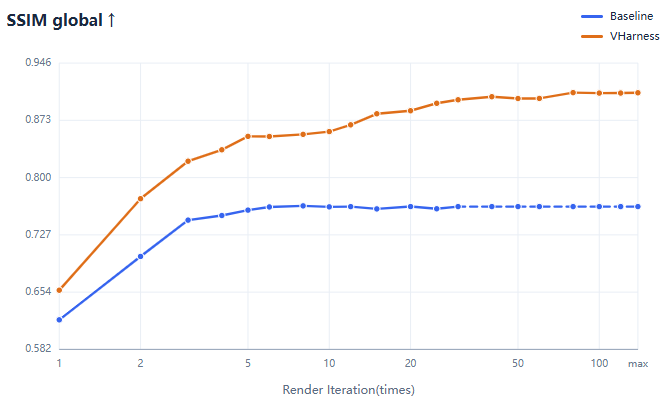}
        \caption{GPT-5.6-sol on global SSIM ($\uparrow$).}
        \label{fig:gpt56_ssim_3d}
    \end{subfigure}

    \caption{\textbf{Cross-model and cross-metric refinement trajectories on Image-to-3D.} We plot DINOv2 (left) and global SSIM (right) across render iterations for GPT-5.5 (top) and GPT-5.6-sol (bottom). Because the \texttt{w/o VHarness} baseline terminates early, the dashed line extends its final score for visual comparison.}
    \label{fig:3d_refinement_dynamics}
\end{figure*}

Figure~\ref{fig:3d_refinement_dynamics} tracks the mean DINOv2 and global SSIM scores on Image-to-3D for GPT-5.5 and GPT-5.6-sol, where dashed horizontal lines extend the \texttt{w/o VHarness} scores after early termination. Across all four plots, \texttt{w/ VHarness} achieves steeper gains at matched early iterations and continues improving well beyond the point where the baseline plateaus. This demonstrates that VHarness achieves higher per-iteration refinement efficiency, extracting larger quality gains per render cycle while effectively converting additional token compute into sustained visual and geometric improvements rather than blind trial-and-error.

\subsection{Qualitative Refinement Trajectories}
\label{app:trajectories}

\begin{figure*}[!t]
    \centering
    \includegraphics[width=0.9\textwidth]{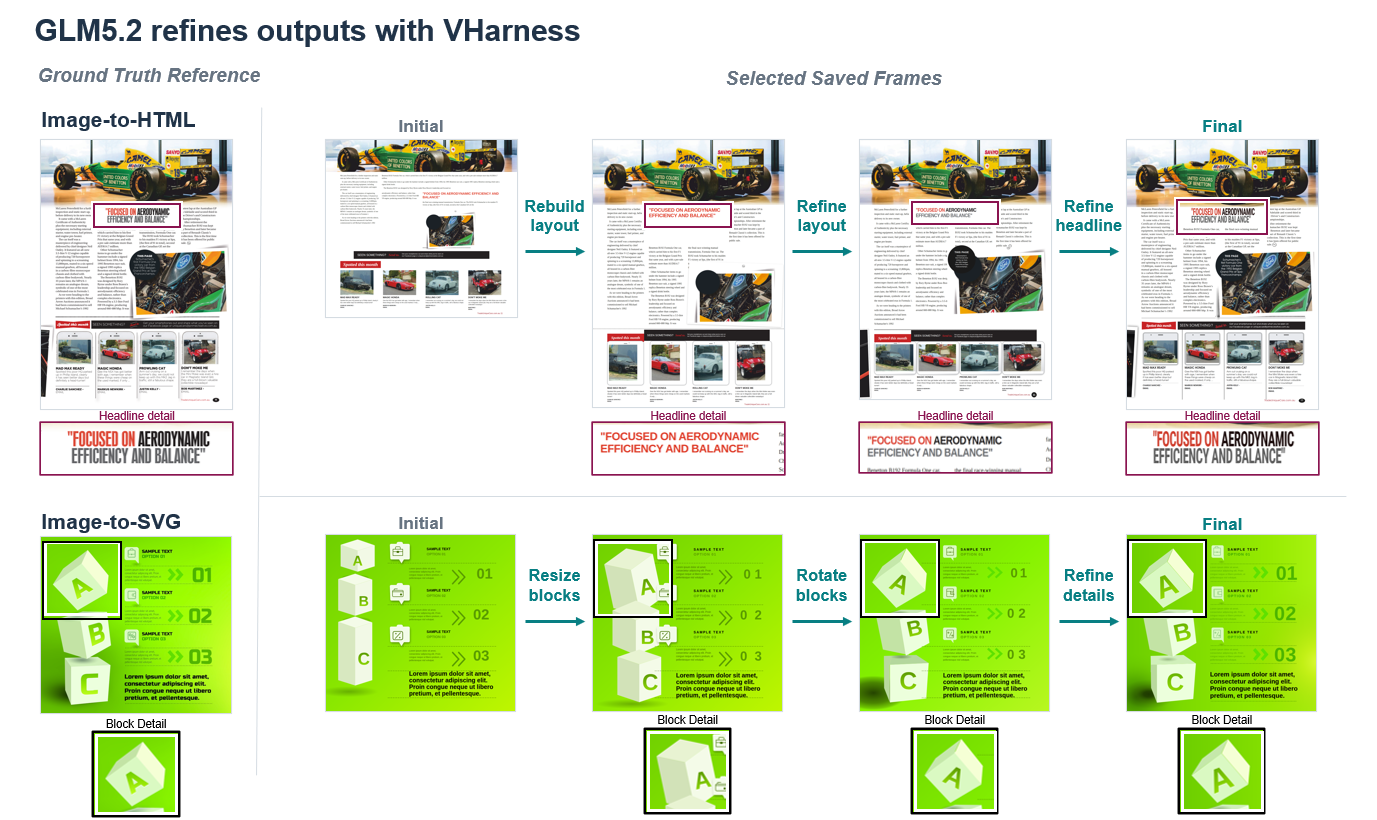}
    \caption{Multi-step refinement trajectories of the text-only coding agent GLM-5.2 with VHarness on Image-to-HTML (top) and Image-to-SVG (bottom), showing selected intermediate renders from the initial draft to the final output alongside zoomed-in regional details.}
    \label{fig:appendix_glm52_traj}
\end{figure*}

Figure~\ref{fig:appendix_glm52_traj} visualizes multi-step refinement trajectories of the text-only agent GLM-5.2 with VHarness on Image-to-HTML (Case 079, top) and Image-to-SVG (Case 021, bottom). Moving from the initial draft (Render \#1) to the final output, the red bounding boxes and bottom zoomed-in callouts illustrate how iterative feedback from \texttt{diff}, \texttt{grep}, and \texttt{inspect} guides the agent to progressively fix layout structure, vector geometry, and fine local details.

\end{document}